\documentclass[a4paper]{cas-dc}

\usepackage[numbers]{natbib}

\usepackage{graphicx}
\usepackage{subcaption}
\usepackage{dirtree}
\usepackage{listings}

\def\tsc#1{\csdef{#1}{\textsc{\lowercase{#1}}\xspace}}
\tsc{WGM}
\tsc{QE}

\begin{document}
\let\WriteBookmarks\relax
\def\floatpagepagefraction{1}
\def\textpagefraction{.001}
\let\printorcid\relax    
\shorttitle{StratMed} 

\shortauthors{Xiang Li}  

\title [mode = title]{StratMed: Relevance Stratification between Biomedical Entities for Sparsity on Medication Recommendation}  



%

\author[1]{Xiang Li}
\ead{lixiang_222@stumail.ysu.edu.cn}
\credit{Conceptualization, Methodology, Software, Validation,  Formal analysis, Data Curation, Investigation, Writing – original draft, Writing - Review \& Editing, Visualization}

\author[1,2]{Shunpan Liang}
\ead{liangshunpan@ysu.edu.cn}
\credit{Supervision, Funding acquisition}
\cormark[1]
\cortext[1]{Corresponding author.}

\author[3]{Yulei Hou}
\ead{ylhou@ysu.edu.cn}
\credit{Resources}

\author[4]{Tengfei Ma}
\ead{tfma@hnu.edu.cn}
\credit{Supervision, Writing - Review \& Editing}

\affiliation[1]{
            organization={School of Information Science and Engineering, Yanshan University},
            city={QinHuangDao},
            postcode={066004}, 
            country={China}
}

\affiliation[2]{
            organization={School of Information Science and Engineering, Xinjiang University Of Science \& Technology},
            city={Korla},
            postcode={841000}, 
            country={China}
}

\affiliation[3]{
            organization={School of Mechanical Engineering, Yanshan University},
            city={QinHuangDao},
            postcode={066004}, 
            country={China}
}

\affiliation[4]{
            organization={School of Computer Science and Engineering, Hunan University},
            city={ChangSha},
            postcode={410012}, 
            country={China}
}


\begin{abstract}
With the growing imbalance between limited medical resources and escalating demands, AI-based clinical tasks have become paramount. As a sub-domain, medication recommendation aims to amalgamate longitudinal patient history with medical knowledge, assisting physicians in prescribing safer and more accurate medication combinations. 
Existing works ignore the inherent long-tailed distribution of medical data, have uneven learning strengths for hot and sparse data, and fail to balance safety and accuracy.
To address the above limitations, we propose StratMed, which introduces a stratification strategy that overcomes the long-tailed problem and achieves fuller learning of sparse data. It also utilizes a dual-property network to address the issue of mutual constraints on the safety and accuracy of medication combinations, synergistically enhancing these two properties.
Specifically, we construct a pre-training method using deep learning networks to obtain medication and disease representations.
After that, we design a pyramid-like stratification method based on relevance to strengthen the expressiveness of sparse data.
Based on this relevance, we design two graph structures to express medication safety and precision at the same level to obtain patient representations. 
Finally, the patient's historical clinical information is fitted to generate medication combinations for the current health condition.
We employed the MIMIC-III dataset to evaluate our model against state-of-the-art methods in three aspects comprehensively. Compared to the sub-optimal baseline model, our model reduces safety risk by 15.08\%, improves accuracy by 0.36\%, and reduces training time consumption by 81.66\%.
Our source code is publicly available at: 
\url{https://github.com/lixiang-222/StratMed}
\end{abstract}



\maketitle

\section{Introduction}
In recent years, due to the outbreak of large-scale infectious diseases, there has been a noticeable skew in the supply-demand dynamics of the healthcare system. An increasing number of patients are experiencing exacerbated conditions due to the lack of medical resources. As a result, AI-based clinical support systems, such as medication recommendation \cite{define-1,define-2}, medical image processing \cite{image-1,image-2}, and medical data analysis or classification \cite{analysis-1, classification-1,classification-2}, have garnered broader attention from society. 
Meanwhile, some basic research, such as patient retrieval methods or machine learning methods \cite{reviewer1_1,reviewer1_2,reviewer1_3}, have also made significant breakthroughs in intelligent healthcare, again driving deeper integration of AI in healthcare.

Medication recommendation using technologies such as data analytic\cite{reviewer3_3,machine-1,machine-2}, deep learning\cite{deep-2,deep-3} to recommend appropriate medication combinations for patients based on their clinical history, health status, and medication knowledge. 
A personalized medication recommendation system effectively reduces prescribing errors due to individual fitness issues, avoids unnecessary medication expenses, and speeds up the treatment process, providing patients with more treatment. It also considers matters easily overlooked in many aspects of manual diagnosis and treatment systems, such as the risk of drug-drug interactions. In conclusion, a safe and appropriate recommendation system on medication is essential for the reliability and sustainability of the healthcare system. 
    
Regarding dataset structure, medication recommendation bears similarities to some sequence recommendations \cite{seq1,seq2,seq3} and session recommendations \cite{session1,session2}. These methods predict the label of the next session based on the label(s) of a single/multiple historical session for a specific feature. However, traditional recommendation systems have the following limitations and cannot be directly migrated.
On the one hand, the extensive presence of drug-drug interaction(DDI) \cite{ddi1,ddi2,ddi3} sets medication recommendations apart from conventional recommendation algorithms that predominantly pursue precision. Medication recommendation necessitates the consideration of the safety of the proposed medication combinations.
On the other hand, in traditional recommender systems, the recommendation of the current item (i.e., product) relies on the past user's interaction with the previous items, and the recommendation result mainly depends on the similarity between the same class entities. In contrast, medication recommendation is mainly based on the current diagnoses for the current medications. The outcome of medication recommendation depends more on the relationship between different classes of entities.

Early investigations \cite{earlywork1,ealywork2,ealywork3} in medication recommendation predominantly centered on present health conditions, inadequately accounting for a patient's longitudinal prescription history across multiple consultations. This limitation restricted the capacity of these studies to formulate tailored medication prescriptions for individual patients.
Subsequent research \cite{longitudinal1,longitudinal2,longitudinal3} incorporated parts of techniques from sequential recommendation systems, achieving notable advancements in enhancing the precision of recommendation outcomes. Nevertheless, these studies frequently overlooked the paramount importance of ensuring medication safety.
As research into drug interactions\cite{ddi(mtf1),ddi(mtf2)} has advanced, recent works \cite{safedrug,molerec} have emphasized the pivotal role of molecules in refining medication recommendations. However, these studies have not addressed the long-tail distribution in medical datasets and consistently face the challenge of finding an ideal balance between the safety and accuracy of recommended medication combinations.

\begin{figure}
    \centering
    \includegraphics[width=0.9\linewidth]{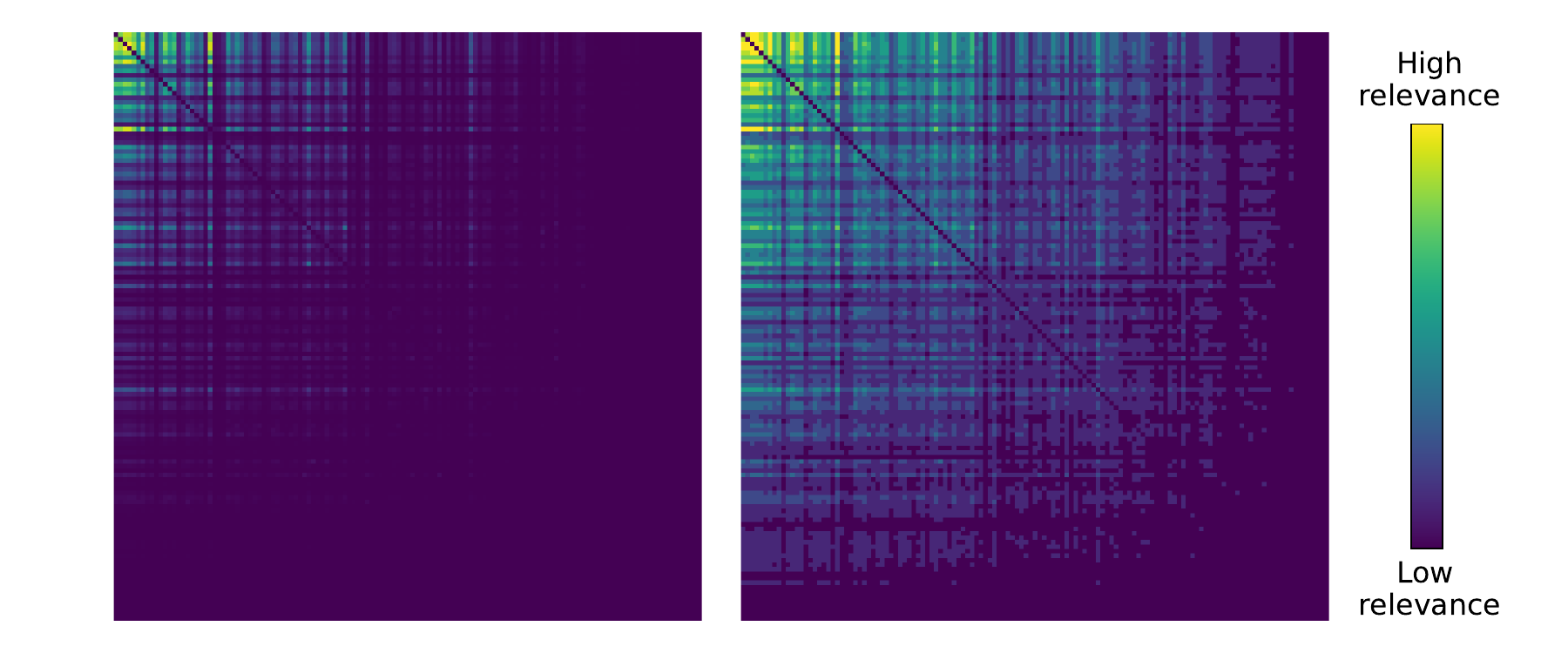}
    \caption{Distribution of inter-medication relationships. The left graph depicts the distribution observed in the original data, whereas the right graph depicts the ideal distribution.}
    \label{fig:data_process}
\end{figure}

In this study, we identify two primary manifestations of inaccuracy in medication recommendations. The first is under-prescription, where the model neglects to suggest necessary medications due to its inadequate representation of certain rare diseases. The second is over-prescription, where the model overly emphasizes common conditions, leading to excessive prescribed medications for these symptoms. The above significant disparity in expression ability between popular and rare data arises from the long-tail distribution present within Electronic Health Records (EHR). 
Therefore, we posit that balancing the long-tail distribution of data can considerably improve the portrayal of a patient's health status, subsequently elevating the efficacy of medication recommendations. The crux of this improvement lies in emphasizing the sparse relationships of entities.
Figure \ref{fig:data_process} presents the distribution of medication correlations within the MIMIC-III dataset. Here, a brighter hue signifies a stronger relevance, while darker regions indicate weaker relevance. As depicted in the left sub-figure, most medications in the original dataset exhibit limited association, leading to a profoundly imbalanced distribution. In contrast, the right sub-figure showcases the distribution after processing with our algorithm, reflecting a more balanced and healthier relationship.
Additionally, existing methods typically prioritize either safety or accuracy, failing to give equal attention to both properties. Such an approach leads to an imbalance between safety and accuracy. We believe that the key to achieve this balance is to integrate both properties on a unified platform and then enhance the overall capability of the recommendation.

Following this line, we introduce the StratMed model, which alleviates the long-tail issue while ensuring a balance between accuracy and safety.
We first designed a pre-training model that integrates diagnosis, procedure, and medication data from visits, using a multi-layer perceptron network to obtain independent embeddings of all data.
To address the issue of sparsity in biomedical entity relationships, we introduce a pyramid-style structure for relationship buckets to store entity relevance. We then employ varying gradient truncation methods tailored to each relationship bucket, emphasizing medium and sparse entity relationships and integrating this entity relevance into subsequent graph networks.
We extract data from a single clinical visit, implement a safety weight graph convolutional network (GCN-SW) to update medication embeddings, and aggregate them in a safety representation. Simultaneously, a mapping features graph convolutional network (GCN-MF) was employed to update the embeddings for diagnosis and procedure. Projecting the diagnoses and procedures into the medication representation space, we aggregate a representation emphasizing medication accuracy. Then we established a comprehensive visit-level representation by harmonizing the safety and accuracy representations. 
Finally, by fitting the longitudinal history of visits from the patient, we get a patient-level representation and recommend a combination of medications for the current health condition.

Compared with the state-of-the-art methods, the stratification method introduced by our model can bolster the learning capacity for sparse data, thereby alleviating the long-tail distribution and enhancing the model's expressive capability. Learning the relationship between medications on one side and the relationship between medications and diseases on the other side, the dual-graph network structure is similar to a balance, which solves the conflict between safety and accuracy that other models fail to deal with and is ahead in terms of comprehensive performance. In addition, our stratification method is based on statistical principles and rules, which eliminates the need to engage in computationally intensive deep learning training, and combined with a pre-training module that can learn key parameters in advance, effectively reduces time costs.

Specifically, the significant contributions of this paper can be summarized as follows: 

\begin{itemize}
    \item We devise a novel stratification method to tackle long-tailed distribution in medical data for medication recommendations. It boosts model performance by emphasizing learning from low and medium-frequency data, effectively countering the long-tailed distribution challenge.
    \item We develop a dual-property (accuracy and safety) representation graph framework. By expressing both pieces of information on the same level, we achieved a balance between medication safety and accuracy.
    \item We conduct extensive experiments on the real dataset MIMIC-III to demonstrate that our proposed method is more effective than existing methods.
\end{itemize}

We list a summary of the various parts of the text below. 
(1) The first section is the Introduction, which introduces the general idea of the paper, its innovations, and the original intention of the work.
(2) The second section is the Related Works, provides an overview of related work, introducing prominent works and current developments in medication recommendation and long-tail problem. 
(3) The third section is the Methods, introduces the core methodology of our model, defining specific problems within the work and providing detailed explanations of the model framework and its implementation. 
(4) The fourth part is the Experiments, which introduces the background related to the experiment, and the experimental methodology and shows the specific experimental results.
(5) The fifth section is the Discussions, which provides an in-depth analysis of the experimental results and a series of supporting experiments.
(6) The last section is the Conclusion, which summarizes the results of the study and concludes with an outlook for future work.

\section{Related Works}
This section will present related work on medication recommendations and long-tail problems.

\subsection{Medication Recommendation}
Medication recommendation is one of the essential parts of bioinformatics, primarily distinguishing into instance-based methods and longitudinal medication recommendation methods.

Instance-based methods emphasize the patient's present health condition. 
LEAP \cite{leap} utilized a multi-instance, multi-label sequence-to-sequence model with content attention for prescriptions. 
Another representative work, DMNC \cite{dmnc} employed a dual memory neural collaborative filtering framework, focusing on medication and patient interactions to produce accurate medication predictions.

In contrast, longitudinal methods provide recommendations based on long-term health records. 
RETAIN \cite{retain} adopted a two-level attention model, highlighting important past visits and extracting significant clinical variables for patient-tailored recommendations. 
GAMENet \cite{gamenet} integrated EHR data with drug-drug interaction knowledge. By creating interconnected EHR-DDI graphs, GAMENet fosters a comprehensive understanding of a patient's medication needs while actively mitigating potential medication conflicts. 
SafeDrug \cite{safedrug} went beyond typical recommendation algorithms by embedding molecular structures, enhancing the predictive quality of recommendations. This molecular-centric approach minimized the risks from DDI. 
MICRON \cite{micron} focused on the importance of the prescription in the patient's last clinical visit and used a residual mechanism to capture changes in a patient's health between two adjacent visits. 
MoleRec \cite{molerec} explored the relationship between a patient's health condition and medication molecular substructures, which can offer more precise medication recommendations tailored to disease specifics. 

Recent advancements in medication recommendation have witnessed the adaptation of methodologies from diverse domains. Some researchers \cite{reviewer3_1} utilized natural language processing techniques to predict accurate medication recommendations by analyzing patient comments and ratings.
CogNet \cite{cognet} treated medication recommendations as a translation model, introducing Transformer into their method, and designed a medication reproduction mechanism, which incorporates those medications that are still applicable from the last visit into the current recommendation. 
DGCL \cite{reviewer3_2} proposed distance detection loss and utilized a graphical contrast learning method for modeling the difference between the output distribution of current and historical records.

Our study seeks to mitigate the challenges of limited healthcare dataset resources, enhancing the sparse relationship among infrequent entities and significantly advancing over existing models.

\subsection{Long-tail Problem}
Within traditional recommendation datasets, long-tail distributions are commonly observed, that a minority of highly favored items attract the most user engagement. Conversely, the majority of items receive minimal user interaction called long-tail items. Long-tail problem \cite{longtail1,longtail2} arose due to the models trained on these datasets tending to excessively specialize in popular items, thus facing difficulties in effectively representing long-tail items.

One prevalent strategy to address the long-tail phenomenon's impact involves resampling. These methods \cite{resampling1,resampling2} entailed introducing redundant instances of long-tail items to achieve a more balanced distribution within the dataset. However, adopting this strategy can inadvertently overfit the resampled items, negatively affecting the model's overall performance. An alternative approach \cite{regularize1} encompassed adjusting or regularizing the loss associated with training examples. This adjustment encourages the model to allocate more significant attention to long-tail items. Beyond these methods, some researchers explored alternative methodologies, such as transfer learning \cite{transfer1} and graph-based information propagation \cite{graph-longtail1}. There is also recent work \cite{pre&fine1} using a combination of fine-tuning and pre-training to address the problem of long-tailed distributions.

As mentioned above, there are limitations in directly generalizing traditional recommendation ideas that address the long-tail problem to medication recommendations. In this paper, we design a pyramid-like data stratification method to alleviate the long-tail problem in medical data and balance the expressiveness of relationships between entities under different distributions.

\section{Methods}
In this section, we illustrate our proposed framework and the implementation details.

\subsection{Problem Definitions}

\newdefinition{definition}{Problem Definition}
\begin{definition}
\textit{(medical entity)}

We define different kinds of medications as medication entities in the model, and any occurrence of medication in the dataset has its corresponding medication entity, where each medical entity corresponds to an independently numbered such as \(\{ {m_1},{m_2},...\}\) and independently same-dimensioned embedding in the model. Similarly, for diagnoses and procedures, there are diagnosis entities and procedure entities, and the three main types of entities are collectively referred to as medical entities.
\end{definition}

\begin{definition}
\textit{(input and output)}

For a given patient \( H \), we represent their Electronic Health Record (EHR) as \( H = \left\{ v(1), v(2), ..., v(t) \right\} \), where \( v(t) \) is the clinical information associated with the \( t^{th} \) visit. In this study, the types of medical data we consider include diagnoses represented as \( d(t) \in \left\{ 0,1 \right\}^{|D|} \), procedures represented as \( p(t) \in \left\{ 0,1 \right\}^{|P|} \), and medications represented as \( m(t) \in \left\{ 0,1 \right\}^{|M|} \). All of these are presented in the dataset using multi-hot encoding. 
We use \( D \), \( P \), and \( M \) to denote the sets of all diagnoses, procedures, and medications found in the dataset, respectively. The respective number of these sets are represented as \( |D| \), \( |P| \), and \( |M| \). For the visit \( v(t) \) of the patient \( H \), the sets of diagnoses, procedures, and medications present are denoted by \( D(t) \subset D \), \( P(t) \subset P \), and \( M(t) \subset M \), respectively.

For the visit \( v(t) \), we utilize all data from \( H \) (excluding \( M(t) \)) to predict the medication combination \( \hat{M}(t) \) that best aligns with the patient's current health condition. Our model is trained by comparing \( \hat{M}(t) \) with the actual medication combination \( M(t) \) from the dataset.
\end{definition}

\begin{definition}
\textit{(DDI graph)}

Drug-drug interaction (DDI) refers to the phenomenon where one drug influences the effect of another one when both are administered together. This interaction can lead to reduced or harmful therapeutic effects on patients. Thus, understanding and identifying these interactions are crucial when recommending medication combinations.

To capture this vital information, we constructed a Med-Med interaction graph, denoted by \(G^{ddi} = \{M, A^{ddi}\}\). In this graph, the node set \(M\) comprises all medications in the dataset. The edge set \(A^{ddi}\) denotes the known drug interactions, where each edge \(a^{ddi}_{ij} \in \{0,1\}\). Specifically, \(a^{ddi}_{ij} = 1\) signifies a known interaction between medication \(i\) and medication \(j\), implying these drugs should ideally not coexist in the same medication combination. It's worth noting that this DDI graph is consistent and shared across all patients and visits.    
\end{definition}

\subsection{StratMed}
\begin{figure*}
    \centering
    \includegraphics[width=\textwidth]{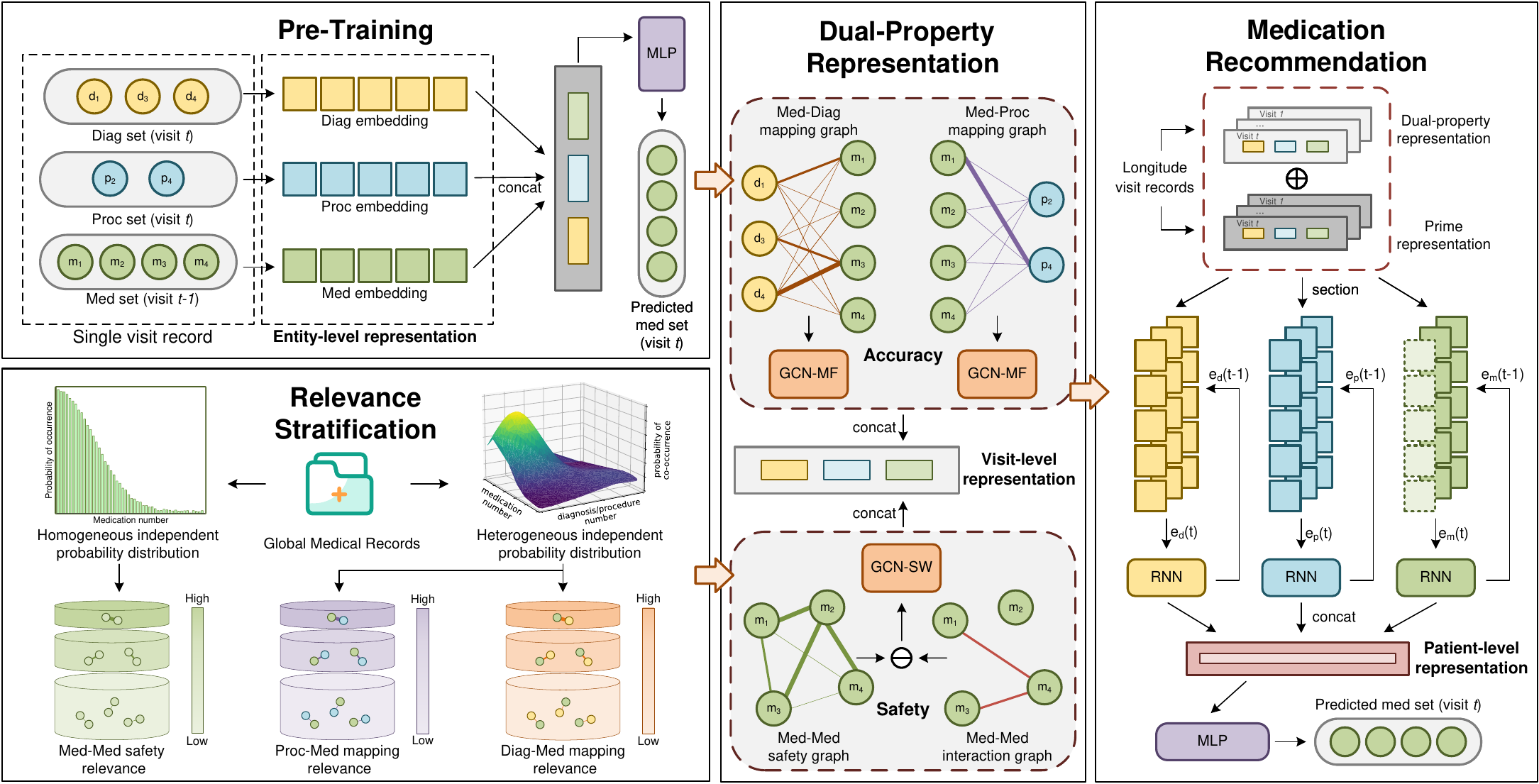}
    \caption{StratMed: We begin with a neural network-based pre-training in the upper left corner to generate entity-level representations $e_d$, $e_p$, and $e_m$ for diagnosis, procedure, and medication. Simultaneously, in the bottom right corner, we stratify the relevance of entity relationships across the global dataset, producing relationships suitable for graph network construction. Following this, in the middle part, we employ a dual-graph network module to update and consolidate the entity information from visit $v(t)$, formulating a visit-level representation $e_v(t)$. In the recommendation phase on the right, we employ a residual approach to the visit representations and integrate multiple longitudinal visit records. We generate the patient-level representation by transitioning to the medication space, \( e^h \). Medications exceeding a specified score threshold will be included in our recommendation.}
    \label{fig:model_graph}
\end{figure*}

As shown in Figure \ref{fig:model_graph}, aggregating from medical entity-level representations to visit-level representations and then to patient-level representations for recommendation, our proposed framework will be bottom-up and is comprised of four major modules:
(1)\textbf{Pre-training} is a prototype of a medication recommendation system that utilizes a multi-layer perceptron network to generate recommendations. Its main purpose is to train the entity-level representation to lay the information foundation for the subsequent messaging module, as well as to speed up the convergence of the primary model.
(2)In the \textbf{Relevance Stratification} phase, we aggregate all training datasets and systematically analyze their frequency distributions. Subsequently, the data is stratified based on this frequency information, and specific enhancements are applied to augment the learning capabilities of the mid to low-frequency data. This approach is employed to alleviate the influence of long-tail distribution and enhance the overall modeling capacity.
(3)In the \textbf{Dual-Property Representation} phase, we designed two different graph network structures that focus on updating and aggregating entity representations in visit and then importing the visit-level representations into subsequent modules. These structures aim to balance safety and accuracy, addressing mutual constraints.
(4)\textbf{Medication recommendations} are used for integration and output, where we will integrate the calendar visits into a patient-level representation and then transform this representation into a medication probability prediction. Medications exceeding a predetermined threshold will be output as the outcome.

\subsubsection{Pre-training}
During the clinical visit, physicians make medication recommendations by referencing the patient's medical history and considering the current diagnosis and procedures. To emulate this practical clinical process, we introduce a pre-training approach to capture the intricacies of biomedical data and obtain entity-level representation.

We encode current diagnoses and procedures as the patient's current health status. Specifically, we first extract the diagnosis/procedure set $d(t) = \{ d_i,d_j,... \} $ and $p(t) = \{ p_k,p_l,... \} $ from the visit $v(t)$, both of which are the multi-hot vectors we have mentioned in the problem definition. We define two learnable embedding tables, $E_d\in R^{|D|\times dim}$ and $E_p\in R^{|P|\times dim}$ corresponding to diagnosis and procedure respectively, where $dim$ is the embedding size. As an entity-level representation shared in the global data, $e_{d_i}/e_{p_k}$ is generated by mapping the diagnosis $d_i$/ procedure $p_k$ to the embedding space. Combine all the diagnosis/procedure embeddings in visit $v(t)$ to get the diagnosis/procedure set presentation $e_d(t)$ and $e_p(t)$.
\begin{equation}
    e_d(t) = d(t)E_d, \quad e_p(t) = p(t)E_p.
\end{equation}

Relying solely on information from a single visit to formulate prescriptions can be limiting. For instance, prescribing medication based purely on present symptoms may be inappropriate when facing an average patient and one with a history of chronic hypertension, and it is essential to factor in potential underlying conditions. Consequently, we incorporate medications from the previous visit \(v(t-1)\) to accommodate a patient's historical health profile. Analogous to the diagnosis, a shared medication embedding table, \(E_m \in R^{|M| \times dim}\), is designed to encode the medication vector in last visit  \( m(t-1) \). It is worth mentioning that if a patient is visiting the hospital for the first time and has no prior medical history, we represent their historical medication records with a zero vector.
\begin{equation}
    e_m(t-1) = m(t-1)E_m.
\end{equation}

Lastly, we concatenate the three vectors of equal length into a single vector thrice the original length. This concatenated vector forms the visit-level representation, \( e'_v(t) \in R^{ 1 \times 3dim}\).
\begin{equation}
    e'_v(t) = \text{CONCAT}(e_d(t), e_p(t), e_m(t-1)).
\end{equation}

The vector \(e'_v(t)\) is then passed through a multi-layer perceptron (MLP) to transform its dimension to \( |M| \), the total number of medications in the dataset, resulting in the final predicted representation \( \hat{m}(t) \), through a $sigmoid$ activation function.
\begin{equation}
    e'^h = \text{MLP}_1(e'_v(t)).
\end{equation}

The output strategy, loss function, and training method are consistent with those in the main model, which we will elaborate on in the following subsections.

As a result of pre-training, we feed the obtained entity-level representations \(e_d, e_p, e_m\) into the subsequent model. Additionally, the visit-level representations \(e'_v\) generated during the training process are used as the source of residual network in the method recommendation phase of the model.

\subsubsection{Relevance Stratification}
In the previous stage, we obtained entity-level representations from single visits. However, when confronted with issues like safety and accuracy of combinatorial recommendation, this representation still lacks generalizability, necessitating a global perspective. Given the sparsity of the dataset, relationships between numerous entities are challenging for the model to capture. 
In this section, we emphasize strengthening the relationships between unpopular biomedical entities, balancing the relevance of various entity relationships within the model, and enhancing the model's representative capability.

Our stratification method groups relationships with similar low-frequency level into a single layer, elevating their position compared to their original distribution while retaining the integrity of the data in the upper layers. Essentially, while the top-tier data remains unchanged, the data in the middle and lower tiers has been uplifted.

Specifically, we tally co-occurrence frequencies and employ the independent distribution of medications and the joint distribution of medication-diagnosis/procedure to construct relevance buckets. We have devised a pyramid-like stratification algorithm tailored to the long-tail distribution of medical data, decomposing diverse relationships into generalized layers and assigning new relevance to the relationships in each layer, which sets the foundation for precise propagation and update on biomedical entities in the subsequent phase.

\noindent\textbf{Safety relevance bucket} 

Upon observation, we found that a high co-occurrence frequency between two medications indicates they typically serve the same condition and have a lower probability of interacting adversely. Conversely, a low co-occurrence frequency suggests the medications are complementary options for the same condition or correspond to entirely different diseases, with a higher likelihood of interaction. Thus, we believe the relevance of safety between medications derives from their co-occurrence frequency.

We design a stratification approach to model medication relevance. Initially, we formulated a medication relationship bucket \(B^{m-m}\), capturing all medication pairings in the global dataset while documenting the co-occurrence frequency for every medication combination. 
\begin{equation}
    \sum^{n_{m-m}}_{i=1}|b^{m-m}_i| = |B^{m-m}| = |M| \times |M|,
    \label{safety relevance equation 1}
\end{equation}
where \( i \) is the specific layer index, \( n_{m-m} \) stands for the total number of layers,  \( |b^{m-m}_i| \) denotes the number of relationships within a specific layer, \( |B^{m-m}| \) represents the total number of relationships in the medication relationship bucket, \( |M| \) is the total count of medications in the global dataset.

Utilizing this co-occurrence data, we infer an isomorphic independent probability distribution of medication relevance. Subsequently, we modulated the excessively steep gradients within this distribution, transitioning towards a relatively flat pyramid-like stratification structure. 

In this medication safety pyramid, the top layer comprises the least yet highly correlated relationships, whereas the foundational layer is characterized by a vast volume of weakly correlated relationships. We set the top-tier data volume as \(q_{m-m}\). Together with the gradient coefficient \( k \), it determines the data distribution within the bucket as well as the quantity within any layer \( i \), denoted as \( b^{m-m}_i \).
\begin{equation}
    |b^{m-m}_i| = q_{m-m} k^{i-1},
    \label{safety relevance equation 2}
\end{equation}
based on equations \ref{safety relevance equation 1} and \ref{safety relevance equation 2}, we can determine the number of layers \( n_{m-m} \) in the stratification:
\begin{equation}
    n_{m-m} = {\log{\frac{\frac{|M| \times |M|}{q_{m-m}-1} }{k-1}}}/{\log k}.
\end{equation}

\noindent\textbf{Mapping relevance bucket}

Unlike conventional recommendation algorithms that rely on homogeneous historical data to predict current data, the distinctive essence of the medication recommendation framework hinges on generating recommendations based on heterogeneous current data. Therefore, enhancing the precision of medication recommendations relies on accurately capturing the mapping relevance between medications and diagnoses/procedures.

Analogous to the relationships among medications, we posit that the mapping relevance between medication and diagnosis/procedure arises from their co-occurrence frequencies. However, the interpretation of lower co-occurrence rates between medication and disease differs from that between medications. 
When the co-occurrence rate is extremely low, it suggests that there's virtually no relationship between the medication and the disease. However, at moderately low frequencies, a certain relationship typically exists, and the relationship appears sparsity due to the rarity of the data. Our stratification strategy aims to amplify the representation of such relationships. To ensure the precision of the desired relationships, during the construction of the heterogeneous relevance buckets \( B^{m-d} \) and \( B^{m-p} \), in addition to establishing a top layer, we introduce a bottom layer. Relevance from the bottom layer represents an extremely low co-occurrence rate on the relationship between medications with diagnoses/procedures and will be erased in the following stratification stages,
\begin{equation}
\begin{gathered}
    \sum^{n_{m-d}}_{j=1}|b^{m-d}_j| = |B^{m-d}| = |M| \times |D| - |\{c_{m-d} < \theta \}|,\\
    \sum^{n_{m-p}}_{l=1}|b^{m-p}_l| = |B^{m-p}| = |M| \times |P| - |\{c_{m-p} < \theta \}|,
\end{gathered}
\label{mapping relevance equation 1}
\end{equation}
where \( |D| \) and \( |P| \) respectively represent the total number of diagnoses and procedures in the dataset, and \( c_{m-d} \) and \( c_{m-p} \) denote the co-occurrence counts between medications and diagnoses/procedures. \( \theta \) is defined as excessively low co-occurrences threshold, set as 0.03\% of the total visit count in this study. And \(|\{c_{m-d} < \theta \}|\) is the amount of data to be set as the bottom layer.

We then apply the stratification method similar to safety relevance, set the data volume of the two top layers \(q_{m-d}\) and \(q_{m-p}\), and combine the gradient coefficient \( k \) to calculate the data volume in each layer,
\begin{equation}
    |b^{m-d}_j| = q_{m-d} k^{j-1}, \quad |b^{m-p}_l| = q_{m-p} k^{l-1}, 
    \label{mapping relevance equation 2}
\end{equation}
based on equations \ref{mapping relevance equation 1} and \ref{mapping relevance equation 2}, we can determine the number of layers \( n_{m-d} \) and \( n_{m-p} \) in the stratification:
\begin{equation}
\begin{gathered}
    n_{m-d} = {\log{\frac{\frac{|M| \times |D| - |\{c_{m-d} < \theta \}|}{q_{m-d}-1} }{k-1}}}/{\log k},\\
    n_{m-p} = {\log{\frac{\frac{|M| \times |P| - |\{c_{m-p} < \theta \}|}{q_{m-p}-1} }{k-1}}}/{\log k}.
\end{gathered}
\end{equation}

\noindent\textbf{Relevance assignment}

Upon reorganizing the three sets of relationships, we achieved a more uniformly distributed stratification with a gentler slope. We assigned relevances to the relationships within these strata to align with the subsequent graph network. For the relevance \( r^{m-m} \in b^{m-m}_i\) of safety relationship within a given layer $i$:
\begin{equation}
    r^{m-m}_i = \frac{i}{n_{m-m}},
\end{equation}
the same method is applied to the mapping relationship,
\begin{equation}
    r^{m-d}_j = \rho_{m-d} \frac{j}{n_{m-d}}, 
    \quad 
    r^{m-p}_l = \rho_{m-p} \frac{l}{n_{m-p}},
\end{equation}
where \( r^{m-d} \in b^{m-d}_j\) and \(r^{m-p} \in b^{m-p}_l\) are the relevance of the relationship in a given layer, the predefined maximum relevance for med-diag and med-proc is represented by \( \rho_{m-d} \) and \( \rho_{m-p} \) respectively.

\subsubsection{Dual-property Representation}
In this section, we aim to represent the safety and accuracy of medications on the same platform. Specifically, by considering each visit as a distinct session, we initially construct a complete graph for medication to represent safety, followed by a bipartite graph for medication-diagnosis/procedure to represent accuracy. Subsequently, we aggregate two representations into visit-level representation.

\noindent\textbf{Safety representation} 

For representation of safety, we first define an undirected complete graph $G^{m-m}(t)= \left \{M(t-1), A^{m-m}\right \}$, where $M(t-1)$ is the set of all medications in visit $v(t-1)$, and \(a^{m-m}\) is a 0-1 adjacency matrix representing the edge set. Each specific node $m \in M(t)$ is associated with corresponding medication embedding $e_m$. 

\begin{figure}
    \centering
    \includegraphics[width=0.75\linewidth]{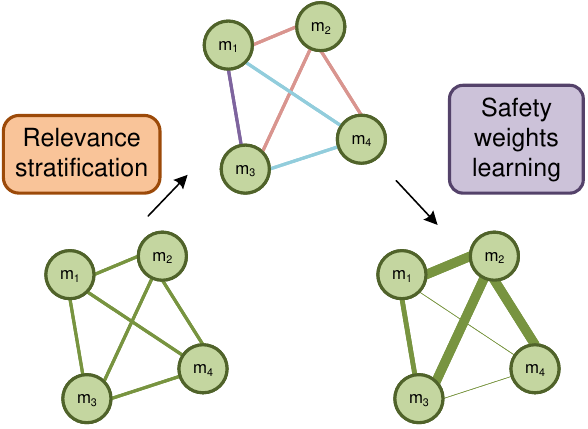}
    \caption{Edge weight change in Med-Med safety graph based on GCN-SW, the different thicknesses of the lines in the edges represent the differences in the weights obtained by the classified edges in training, with thicker edges representing larger weights (stronger relevance) and thinner edges representing smaller weights (weaker relevance or exist interaction side effects).}
    \label{GCN-SW}
\end{figure}

Figure \ref{GCN-SW} illustrates the weight change on the edges of the Med-Med safety graph during the whole process of the model. 
In GCN-SW, we establish a weight embedding vector \( E_{a^{m-m}} \in R^{|A^{m-m}|\times 1} \). Each 1D embedding within this vector maps to a relationship of a specific layer. We use the relevance corresponding to the relationship as the initial weight $w'^{m-m}$ for the edges in the graph.
\begin{equation}
    w'^{m-m}_{ij}=\{b_{l_1}^{m-m}:a^{m-m}_{ij} \in b_{l_1}^{m-m}\} E_{a^{m-m}} = r^{m-m}_{l_1},
\end{equation}
where \(i\) and \(j\) denote the indices of two medications, \(a^{m-m}_{ij}\) signifies the relationship between the two medications, \( b_{l_1}^{m-m} \) denotes the layer to which \( a^{m-m}_{ij} \) belongs in the relationship bucket, and \(r^{m-m}_{l_1}\) denotes the relevance generated for that layer of relationship during the stratification phase.
                
Additionally, some medications have the DDI and cannot be used together. We also need to consider this unsafe relationship when modeling the medication combination graph, so we use the Med-Med interaction graph and combine it with the Med-Med safety graph.
\begin{equation}
    w^{m-m}_{ij} = w'^{m-m}_{ij} - \lambda a^{ddi}_{ij},
\end{equation}
where $a^{ddi}_{ij}$ is the interaction relationship mentioned in the previous section, $\lambda$ is the learnable parameter, and $w^{m-m}_{ij}$ is the final weights of the edge between medications.

Next, we use a GCN-based graph network to pass the medication graph as follows:
\begin{gather}
    h^{m-m}_{ij}=\text{MESSAGE}_1(e_{m_i}^{t-1},e_{m_j}^{t-1};w^{m-m}_{ij}),\\
    e_{m_j}^t=\text{UPDATE}_1(e_{m_j}^{t-1},h^{m-m}_{ij} | m_i\in M(t-1)),
\end{gather}
where $h^{m-m}_{ij}$ denotes the message vector passing from $m_i$ to $m_j$ , and $e_{m}^{t-1}$ denotes the embedding representation of medication obtained after the visit $v(t-1)$ trained. During the training, the message pass function $\text{MESSAGE}_1(\cdot)$ and the node update function $\text{UPDATE}_1(\cdot)$ are used to update the representation of each entity.

Using function $\text{AGGREGATION}_1(\cdot)$, we aggregate the updated medication nodes from this iteration to form a medication set representation, \( e_v^m(t-1) \), that reflects safety:
\begin{equation}
    e_m(t-1) = \text{AGGREGATION}_1(e_{m_i}^t | m_i\in M(t-1)).
\end{equation}

\noindent\textbf{Accuracy representation}

For the representation of accuracy, we define two bipartite fully connected graphs $G^{m-d}(t) =  \{D(t), M(t-1), A^{m-d}\}$ and $G^{m-p}(t) =  \{P(t), M(t-1), A^{m-p} \}$. In the bipartite graph, different types of entities are linked to each other. Each edge is a directional edge that starts with a medication and ends with a diagnosis/procedure, where $D(t)/P(t)$ is the set of all medications in visit $v(t)$, and \(A^{m-d}/A^{m-p}\) is a 0-1 adjacency matrix representing the edge set, the node in $D(t)/P(t)$ is associated with corresponding diagnosis/procedure embedding $e_d/e_p$.

\begin{figure}
    \centering
    \includegraphics[width=0.75\linewidth]{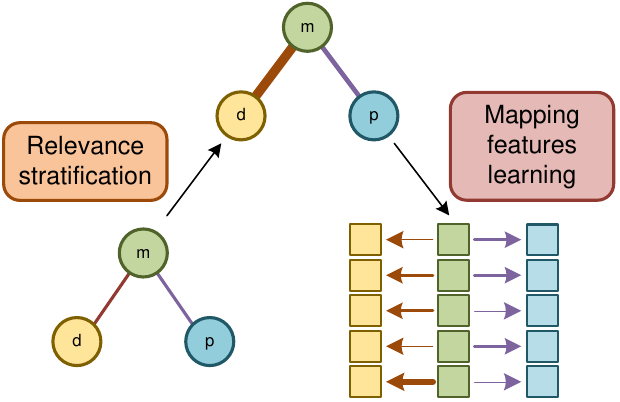}
    \caption{Edge weight change in Med-Diag/Proc mapping graph based on GCN-MF, the different thicknesses of the edges between entities in the upper part of the figure have the same meaning as expressed in GCN-SW. There are also multiple edges with different thicknesses between the features in the lower right of the figure, representing the differences in the weights in the features of the mapping relationships between the different categories of entities.}
    \label{GCN-MF}
\end{figure}

In order to enhance the symptomatic ability of medication and overcome the different embedding space of various types of entities, as shown in Figure\ref{GCN-MF}, we applied a graph network based on mapping features(GCN-MF). We build an embedding table $E_{a^{m-d}}\in R^{|A^{m-d}|\times dim}/ E_{a^{m-p}}\in R^{|A^{m-p}|\times dim}$ as the attributes of $a^{m-d}$ and $a^{m-p}$ to map the medication-diagnosis/procedure features, and each row of the table independently maps the type of layer correlation in the relevance bucket.

Similar to the method for the Med-Med safety graph, we initialize and update the two bipartite graphs separately:
\begin{equation}
\begin{gathered}
    w^{m-d}_{ij}=\{l:a^{m-d}_{ij} \in l_2\} E_{a^{m-d}} = r^{m-d}_{l_2},\\
    w^{m-p}_{ik}=\{l:a^{m-p}_{ik} \in l_3\} E_{a^{m-p}} = r^{m-p}_{l_3},
\end{gathered}
\end{equation}
where \(i, j, k\) denote the indices of medication, diagnosis, and procedure, we eventually obtained \( w^{m-d}_{ij} \in R^{1 \times dim} \) and \( w^{m-p}_{ik} \in R^{1 \times dim} \). And then we pass messages and update the mapping bipartite graph:
\begin{equation}
\begin{gathered}
    h^{m-d}_{ij} = \text{MESSAGE}_2(e_{m_i}^{t},e_{d_j}^{t-1};w^{m-d}_{ij}),\\
    h^{m-p}_{ik} = \text{MESSAGE}_3(e_{m_i}^{t},e_{p_k}^{t-1};w^{m-k}_{ik}),
\end{gathered}
\end{equation}

\begin{equation}
\begin{gathered}
    e_{d_j}^t = \text{UPDATE}_2(e_{d_j}^{t-1},h^{m-d}_{ij} | m_i\in M(t-1)),\\
    e_{p_k}^t = \text{UPDATE}_3(e_{p_k}^{t-1},h^{m-p}_{ij} | m_i\in M(t-1)).
\end{gathered}
\end{equation}

After utilizing the embedding space of medication to depict both diagnoses and procedures, we eventually aggregate them to formulate the diagnosis set representation and procedure set representation, which capture the accuracy of the recommended result:
\begin{equation}
\begin{gathered}
    e_d(t) = \text{AGGREGATION}_2(e_{d_j}^t | d_j\in D(t)),\\
    e_p(t) = \text{AGGREGATION}_3(e_{p_k}^t | p_k\in P(t)).
\end{gathered}
\end{equation}

It is worth mentioning that in the message passing and updating phase, we use a graph network based on GCN to process the graph structure data. GCN \cite{gcn} is a graph network that considers both the characteristics of itself and the characteristics of its neighbor. In GCN-SW, we follow the traditional method. In GCN-MF, we modify it so that the network does not consider itself and focuses on neighbors' information only.

The reason for this modification comes from the unique needs of our model. We intend to express diag/proc based on the relationship between med and diag/proc, so the surrounding neighbors of each node are all heterogeneous nodes, and we need to focus more on the neighboring nodes, avoiding incorporating the node's information as much as possible. For the same reason, in GCN-MF, we use only one layer for our graph network, and a multi-layer graph network may introduce considerations of itself that do not fit the original purpose of our task. With these adjustments to the graph network, we could effectively model and capture the features of the graph data provided in the upstream work.

\noindent\textbf{Integrated visit presentation}

After processing through the graph neural network, we integrate the two accuracy representations (diagnosis and procedure) and one safety representation (medication) to obtain a comprehensive visit-level representation.
\begin{equation}
    e_v(t) = \text{CONCAT}(e_d(t),e_p(t),e_m(t-1)).
\end{equation}

\subsubsection{Medication Recommendation}
As the final part of the model, this subsection aims to adapt the long-term health status by integrating the patient's visit records and recommending relevant medication combinations for the current health representation.

Using \(v(t)\) as an example, we employ a residual approach on visit-level representation. Specifically, we sum the prime representation \( e'_v(t) \) obtained from pre-training and the dual-property representation \( e_v(t) \) from the graph network to yield \( e''_v(t) \). Subsequently, this is divided into three segments based on their information source (diagnosis, procedure, medication): \( e''_d(t) \), \( e''_p(t) \), and \( e''_m(t) \). 
\begin{gather}
    e''_v(t) = e_v(t) + e'_v(t),\\
    e''_d(t),e''_p(t),e''_m(t-1) = \text{SEGMENT}(e''_v(t)),
\end{gather}

To accommodate the long-term health status of patients, we employed three RNNs to model historical visits, producing three segmented representations at the patient level.
\begin{equation}
\begin{gathered}
    e^h_d = \text{RNN}_d(e''_d(t),...,e''_d(1)),\\
    e^h_p = \text{RNN}_p(e''_p(t),...,e''_p(1)),\\
    e^h_m = \text{RNN}_m(e''_m(t-1),...,e''_m(1)).
\end{gathered}
\end{equation}

Specifically, our study employs gated recurrent units (GRUs) \cite{gru} as the core structure of recurrent neural networks. We choose GRU because it has more powerful modeling capabilities compared to traditional RNNs, and is better able to capture temporal dependencies in time-series data while avoiding some of the potential problems in long-short-term memory networks (LSTMs). The relatively more uncomplicated structure of the GRU and its faster training speed make it more suitable for our task. Our input to GRU is the embedding of three kinds of information (diag/proc/med) of the patient's complete visit sequence, and the output of the model serves as the final representation of that kind of information for this visit, which carries the model's learning and summarization of the patient's entire visit history.

Next, we merge the three RNN-passed results and pass them through a MLP to get the final patient representation $e^h$, representing the probability of each medication in that prescription. We select those items whose values are more significant than a predefined threshold $\delta$ to obtain the recommended combination of medications.
\begin{gather}
    e^h = \text{MLP}_2(\text{CONCAT}(e^h_d,e^h_p,e^h_p),\\
    \hat{m}_i = 
    \begin{cases}
        1 ,& \text{if}\quad e^h_i \ge \delta  \\
        0 ,& \text{if}\quad e^h_i < \delta 
    \end{cases}
    .
\end{gather}
where $\hat{m}$ is a multi-hot vector representing the predicted set of medications, the subscript $i$ denotes the $i^{th}$ entry of the vector, $\hat{m}_i=1$ means the medication $m_i$ is recommended, or it is not recommended.

\subsection{Model Training}
In the training session, we optimize all learnable parameters: entity embedding table $E_d$, $E_p$, $E_m$, relationship embedding table $E_{a^{m-m}}$, $E_{a^{m-d}}$, $E_{a^{m-p}}$, and DDI coefficient $\lambda$, and a few RNN, MLP layers. In the inference phase, the model works in the same pipeline as training. Additionally, the loss computation method we used is consistent with that of our benchmark algorithm\cite{safedrug}.

\noindent\textbf{Multiple Loss Function} 

We treat the recommendation task as a multi-label binary classification task. We use two loss functions, binary cross entropy loss $L_{bce}$, and multi-label margin loss $L_{multi}$. The specific calculations are as follows,
\begin{gather}
    L_{bce} = -\sum_{i=1}^{|M|}{m_i}\log({\hat{m}_i})+(1-{m_i})\log (1-{\hat{m}_i}),\\
    L_{multi} = \sum_{i,j:{m_i}=1,{m_j}=0} \frac{\max(0,1-({\hat{m}_i}-{\hat{m}_j})}{|M|},
\end{gather}
where $m_i$ denotes the real value of medication $i$ in the prescription and $\hat{m}_{i}$ denotes the predicted value in our model.

Then we define the DDI loss by summing the probability values of all pairs of medications in which DDI occurs in this recommendation, 
\begin{equation}
    L_{ddi} = \sum_{i=1}^{|M|}\sum_{j=1}^{|M|} a^{ddi}_{ij}\cdot{\hat{m}_i}\cdot {\hat{m}_j}.
\end{equation}

\noindent\textbf{Combined Controllable Loss Function} 

DDI can be present even in prescriptions written by specialized physicians, and excessive focus on DDI will only affect the accuracy and effectiveness of prescribing. So, in the loss combination phase, we balance the importance of each loss component in the model by weighted summation,
\begin{equation}
    Loss = \beta (\gamma L_{bce}+(1-\gamma)L_{multi})+(1-\beta )L_{ddi},
\end{equation}
where $\gamma$ and $\beta$ are hyperparameters.

\section{Experiments}

In this section, we display a series of comparative experiment results between the baseline model and our model and explain the experimental setup and related evaluation metrics in detail.

\subsection{Setup Protocol}
We will begin by introducing the specific experimental environment, the configurations and parameters of the model, and the sampling methodology for the testing phase.

\subsubsection{Experimental Environment}
All experiments are conducted on an Ubuntu 22.04 machine with 30GB memory, 12 CPUs, and a 24GB NVIDIA RTX3090 GPU with pytorch 2.0.0 and CUDA 11.7.

\subsubsection{Configuration and Parameter}
For embedding tables of entities $E_d$, $E_p$, and $E_m$, we use $dim$ = 64 as the embedding size and initialized in pre-training to a uniform distribution from -0.1 to 0.1. For each graph net, we choose a 1-layer GCN with no hidden embedding, and as mentioned before, we set the maximum relevance for med-diag and med-proc $\rho_{m-d}$ = $\rho_{m-p}$ = 0.8. For multi-layer perceptron networks, we set $\text{MLP}_1$ as one linear layer plus basic ReLU activation function and $\text{MLP}_2$ as one linear layer plus Sigmoid activation function, and the dropout between each linear layer is 0.5. For recurrent neural networks, $\text{RNN}_d$, $\text{RNN}_p$, and $\text{RNN}_e$, we use a gated recurrent unit (GRU) with 64 hidden units. For the loss function, we use the same hyperparameters for the training, testing, and validation processes, where threshold $\delta$ = 0.5, $\beta$ = 0.95, $kp$ = 0.05, and acceptance rate $\gamma$ is selected 0.06. We trained 15 epochs in each training and pre-training phase, and parameters are trained on Adam optimizer with learning rate $lr$ = 0.0005, regularization factor $R$ = 0.05.

\subsubsection{Sampling Approach}

Because publicly accessible Electronic Health Record (EHR) data are scarce, we employ bootstrapping sampling during the phase, as suggested following \cite{safedrug}. This method is well-suited for situations involving a paucity of samples, as discussed in references \cite{testsample1} and \cite{testsample2}.

\subsection{Dataset}

MIMIC-III(Medical Information Mart for Intensive Care III) is a widely used medical dataset for researching and analyzing clinical data from intensive care units, such as clinical records, physiological monitoring, lab results, medication records, and tens of thousands of medical images of ICU patients. We divided the dataset into training sets, validation sets, and test sets according to the ratio of 2/3-1/6-1/6 in our extracted data, which contained 15,032 visits, 6,350 patients, 131 medications, 1,958 diagnoses, and 1,430 procedures.

\begin{figure}
    \centering
    \includegraphics[width=0.9\linewidth]{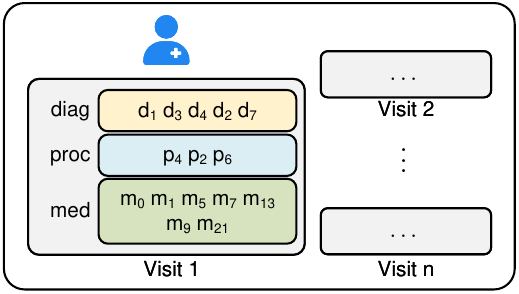}
    \caption{Patient data structure sample after data preprocessing}
    \label{fig:data_sample}
\end{figure}

In order to seamlessly integrate with potential future medical datasets, we have followed a straightforward and efficient data processing method, as earlier work \cite{safedrug}. We sequentially assign numbers to the diagnoses, procedures, and specific medication names in a patient's multiple medical visits and organize them in a standardized structure as Figure \ref{fig:data_sample}. Each patient is independent of the other, and all contain one or more visits, with each visit containing an unordered diag/proc/med collection.

\subsection{Evaluation Metrics}
We analyzed and evaluated the performance of our method in depth with the help of four key metrics, i.e., Jaccard, DDI rate, F1, and PRAUC. In the following, we will describe in detail how each metric is calculated and how it is applied in our study.

\textbf{Jaccard} (Jaccard Similarity Score) is a simple and widely used metric to measure the similarity of two sets. It is calculated by dividing the size of the intersection of two sets by the size of their concurrent sets. In the context of medication recommendation, a higher Jaccard represents a higher degree of accuracy as the predicted prescription is closer to the real label,

\begin{gather}
    Jaccard(t) = \frac{|\{i:\hat{m}_i=1\}|\cap|\{i:{m}_i=1\}|}{|\{i:\hat{m}_i=1\}|\cup |\{i:{m}_i=1\}|}, \\
    Jaccard = \frac{1}{N_h}\sum_{t=1}^{N_h} Jaccard(t),
\end{gather}

where $\hat{m}_i$ represents the multi-hot vector of the predicted outcome, $m_i$ represents the real label, $Jaccard(t)$ represents the evaluation result at visit $t$, and $N_h$ represents the total number of visits for patient $h$.

\textbf{DDI} (Drug-Drug-Interaction Rate) is an essential metric in medication recommendation and refers to the rate of drug interactions present in a drug combination. A lower DDI rate ensures that recommended medication combinations are safer and more effective in clinical practice,
\begin{equation}
    DDI = \frac{\sum_{i=1}^{N}\sum_{k,l\in \{j:\hat{m}_j(t)=1\}}1\{a^{ddi}_{kl}=1\}}{\sum_{i=1}^{N}\sum_{k,l\in \{ j:m_j(t)=1\}}1 },
\end{equation}
where $N_h$ denotes the total number of visits for patient $h$, $m(t)$ and $\hat{m}(t)$ denote the real and predicted multi-label at the visit $t$, $m_j(t)$ denotes the $j^{th}$ entry of $m(t)$, $a^{ddi}$ is the prior DDI relation matrix and $1$ is an indicator function which returns 1 when $a^{ddi}$ = 1, otherwise 0.

\textbf{F1} (F1-score) is an evaluation metric that combines the precision and recall of a model. A high F1 score in medication recommendation indicates that the model balances accuracy and recall to identify appropriate medications for patients while maximizing the capture of possible medication choices,
\begin{gather}
    Precision(t) = \frac{|\{i:\hat{m}_i=1 \}\cap \{i:m_i=1\}|}{|\{i:\hat{m}_i=1\}|},\\
    Recall(t) = \frac{|\{i:\hat{m}_i=1 \}\cap \{i:m_i=1\}|}{|\{i:m_i=1\}|},\\
    F1(t) = \frac{2}{\frac{1}{Precision(t)}+\frac{1}{Recall(t)}},\\
    F1 = \frac{1}{N_h}\sum_{i=1}^{N_h}F1(i),
\end{gather}
where $N_h$ represents the total number of visits for patient $h$.

\textbf{PRAUC} (Precision-Recall Area Under Curve) is the area under the Precision-Recall curve and is used to evaluate the performance of machine learning. A high PRAUC score indicates that the model can maintain high precision at high recall rates, i.e., recommend the proper medication while maintaining a low error rate,
\begin{gather}
    PRAUC(t) = \sum_{k=1}^{|M|} Precision_{k}(t)\triangle Recall_k(t),\\
    \triangle Recall_k(t) = Recall_k(t) - Recall_{k-1}(t),
\end{gather}
where $|M|$ denotes the number of medications, and $k$ is the rank in the sequence of the retrieved medications, and $Precision_{k}(t)$ represents the precision at cut-of $k$ in the ordered retrieval list and $\triangle Recall_k(t)$ denotes the change of recall from medication $k-1$ to $k$. We averaged the PRAUC across all of the patient's visits to obtain the final result,
\begin{equation}
    PRAUC = \frac{1}{N_h}\sum_{i=1}^{N_h}PRAUC(t),
\end{equation}
where $N_h$ represents the total number of visits for patient $h$.

\textbf{Avg.\# of Drugs} (Average number of drugs) refers to the average number of medications included in each recommendation. The significance of this metric is to assess the complexity of the combination of medications provided by the recommender system. A higher metric means that each recommended regimen contains more medications, which may increase the complexity of medication and the risk of adverse effects for patients. Conversely, a lower indicator means that the medication combinations may be easier to manage and reduce unnecessary medication use,
\begin{equation}
    Avg.\# of Drugs = \frac{1}{N_h}\sum_{i=1}^{N_h}|\hat{M}(i)|,
\end{equation}
where $N_h$ represents the total number of visits for patient $h$ and $|\hat{M}(i)|)$ denotes the number of predicted medications in visit $i$ of patient $h$.

\subsection{Baselines}
We chose the following representative state-of-the-art methods as a baseline to test our model:
                
\textbf{LR} (Logistic Regression) is a linear model for binary classification, mapped to a probability space by a linear combination of feature weights, suitable for probabilistic prediction and classification.

\textbf{ECC} \cite{ecc} (Ensemble Classifier Chain) enhances prediction by linking multiple classifiers, each using the output of the previous one. It is commonly used in multi-label classification tasks to improve model performance.

\textbf{RETAIN} \cite{retain} is an attention model for sequence data. It can consider both temporal and feature information for disease prediction and diagnosis. The model dynamically learns important clinical events and features by encoding a patient's clinical history to provide personalized predictions and diagnoses.
                
\textbf{DMNC} \cite{dmnc} is a medical recommendation approach focusing on the association between diseases and medications. DMNC employs a co-attention mechanism that considers information about diseases and medications to capture their complex relationship better.
                
\textbf{GAMENet} \cite{gamenet} is a medical recommendation model that combines a graph neural network and a memory network. GAMENet utilizes graph structures and memory mechanisms to capture correlation information and temporal dependencies in medical sequential data to improve prediction accuracy.
                
\textbf{SafeDrug} \cite{safedrug} looks at the risk of adverse drug interactions. The model considers chemical and biological information between medications and the patient's health status to provide safer medication combination recommendations.
                
\textbf{MICRON} \cite{micron} emphasizes individualized medication combinations. MICRON updates historical medication combinations in the presence of new symptoms to maximize therapeutic efficacy and minimize adverse effects.

\textbf{COGNet} \cite{cognet} is a medication recommendation based on the Transformer architecture. It utilizes a translation model to deduce medications from diseases. Simultaneously, it introduces a copying mechanism to incorporate helpful medications from previous medication combinations into the current recommendation sequence.
                
\textbf{MoleRec} \cite{molerec} predicts the effects and interactions of medications more accurately by modeling the association of specific molecular substructures within a medication molecule with a patient's health condition.
    
\subsection{Performance Comparison}

\begin{table*}[width=0.9\textwidth,cols=4,pos=h]
  \caption{The performance of each model on the MIMIC-III test set in terms of accuracy and safety was evaluated using five evaluation metrics: Jaccard, DDI rate, F1-score, PRAUC, and Average number of drugs. The best and the runner-up results are highlighted in bold and underlined respectively under t-tests, at the level of 95\% confidence level.}
  \begin{tabular*}{\tblwidth}{@{} LLLLLL@{} }
   \toprule
    method  & Jaccard$\uparrow$     & DDI$\downarrow $      & F1-score$\uparrow$    & PRAUC$\uparrow$       & Avg.$\#$ of Drugs\\
   \midrule
    LR                  & 0.4924 $\pm$ 0.0027 & 0.0830 $\pm$ 0.0007 & 0.6496 $\pm$ 0.0025   & 0.7548 $\pm$ 0.0031   & 16.0489 $\pm$ 0.1420 \\
    ECC \cite{ecc}      & 0.4856 $\pm$ 0.0031 & 0.0817 $\pm$ 0.0007 & 0.6438 $\pm$ 0.0028   & 0.7590 $\pm$ 0.0026   & 16.2578 $\pm$ 0.0992 \\
    \midrule
    RETAIN \cite{retain}& 0.4871 $\pm$ 0.0038 & 0.0879 $\pm$ 0.0012 & 0.6473 $\pm$ 0.0033 & 0.7600 $\pm$ 0.0035 & 19.4222 $\pm$ 0.1682 \\ 
    LEAP \cite{leap}    & 0.4526 $\pm$ 0.0035 & 0.0762 $\pm$ 0.0006 & 0.6147 $\pm$ 0.0036 & 0.6555 $\pm$ 0.0035 & 18.6240 $\pm$ 0.0680 \\
    DMNC \cite{dmnc}    & 0.4563 $\pm$ 0.0031 & 0.0795 $\pm$ 0.0010 & 0.6168 $\pm$ 0.0032 & 0.6731 $\pm$ 0.0024 & 20.0000 $\pm$ 0.0000 \\
    \midrule
    GAMENet \cite{gamenet}      & 0.4994 $\pm$ 0.0033   & 0.0890 $\pm$ 0.0005 & 0.6560 $\pm$ 0.0031 & 0.7656 $\pm$ 0.0046 & 27.7703 $\pm$ 0.1726\\
    SafeDrug \cite{safedrug}    & 0.5126 $\pm$ 0.0027   & \textbf{0.0587 $\pm$ 0.0005} & 0.6692 $\pm$ 0.0025 & 0.7660 $\pm$ 0.0022 & 18.9837 $\pm$ 0.1187\\
    MICRON \cite{micron}        & 0.5219 $\pm$ 0.0021   & 0.0727 $\pm$ 0.0009 & 0.6761 $\pm$ 0.0018 & 0.7489 $\pm$ 0.0034 & 19.2505 $\pm$ 0.2618\\
    COGNet \cite{cognet}        & \underline{0.5316 ± 0.0020} & 0.0858 ± 0.0008 & 0.6644 ± 0.0018 & 0.7707 ±  0.0021 & 27.6279 ± 0.0802\\
    MoleRec \cite{molerec}      & 0.5301 ± 0.0025   & 0.0756 ± 0.0006 & \underline{0.6841 ± 0.0022} & \underline{0.7748 ± 0.0022} & 22.2239 ± 0.1661\\
    \midrule
    StratMed & \textbf{0.5321 $\pm$ 0.0035}  & \underline{0.0642 $\pm$ 0.0005} & \textbf{0.6861 $\pm$ 0.0034} & \textbf{0.7779 $\pm$ 0.0043} & 20.5318 $\pm$ 0.1681\\
   \bottomrule
  \end{tabular*}
  \label{tab:comparison with baseline}
\end{table*}

In this subsection, we use the above baseline to compare with our model regarding safety and accuracy from multiple perspectives. To ensure the effectiveness of the baseline, for models that provide available test files, we directly test using their provided files. For models where the test files are unavailable, we retrain and test them using the optimal parameters mentioned in their papers. The comparison results are shown in Table \ref{tab:comparison with baseline}.

As traditional machine learning-based methods, LR and ECC performed slightly worse with lower accuracy than the other models. They failed to avoid higher DDI rates even though they could prescribe smaller medication packages. For LEAP and DMNC, which are based on sequence generation, even though they use deep learning methods, their overall effectiveness is lower than that of traditional models, which suggests that generative models are unsuitable for medication recommendation. RETAIN is a medication recommendation model migrated from sequence modeling, and its work fails to take into account the relationships between medications, leading to higher DDI rates.

\begin{table}
  \caption{The performance of recent excellent models in training and inference efficiency.}
  \begin{tabular*}{\tblwidth}{@{} LLLLL@{} }
   \toprule
    method & \makecell{Convergence\\ Epoch}  & \makecell{Training Time\\/Epoch(s)} & \makecell{Total Training\\Time(s)} & \makecell{Inference\\ Time(s)} \\
   \midrule
    GAMENet & 39 & 45.31 & 1767.09 & 19.27\\
    SafeDrug & 54 & 38.32 & 2069.28 & 20.15\\
    MICRON & 40 & 17.48 & 699.20 & 14.48\\ 
    COGNet & 103 & 38.85 & 4001.55 & 142.91\\
    MoleRec & 25 & 249.32 &	6233.00 & 32.10\\
    \midrule
    pre-training & 11 & 31.76 & 349.30 & -\\
    stratification & - & - & - & 4.12\\
    \makecell{StratMed \\$w/o$ P+S} & 6 & 132.26 & 793.56 & 17.73\\
    StratMed & - & - & 1160.65 & 21.85\\
   \bottomrule
  \end{tabular*}
  \label{tab:time}
\end{table}

In recent years, with the increasing attention on medication recommendation, this field has made significant progress. Therefore, in the following, we will provide a more comprehensive comparison of outstanding experiments conducted over the past four years.
As these outstanding algorithms are all based on deep learning, the efficiency and effectiveness of the models are equally important. Therefore, we conducted a horizontal comparison with these baselines regarding time expenditure. We first recorded the average training time for each epoch during the model's training phase and the epoch at which each model achieved its optimal result as the convergence epoch. Then, we obtain the total training time. Subsequently, we recorded the time required for all models to perform one round of inference on the test set, where each round of inference is roughly 2,000 recommendations. Table \ref{tab:time} illustrates the performance of all models in terms of efficiency. Our model comprises three components: pre-training, stratification, and the StratMed $w/o$ P+S, where StratMed $w/o$ P+S represents the remaining modules that do not include the pre-training and stratification module. Since pre-training does not directly involve inference, it does not include inference time. The stratification algorithm, on the other hand, is not part of the training process and is only computed once before training. Thus, it only consists of inference time. The total training time encompasses the training time of pre-training, the inference time of the stratification module, and the training time of the StratMed $w/o$ P+S. The inference time includes the inference time of the stratification module and the StratMed $w/o$ P+S.

\begin{figure}
    \centering
    \includegraphics[width=0.9\linewidth]{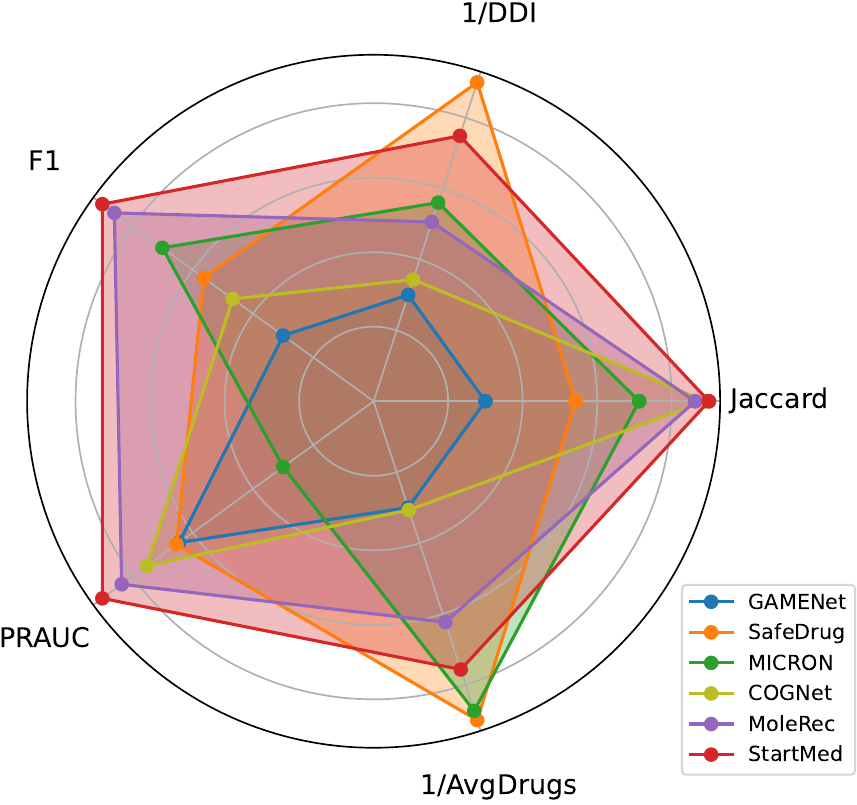}
    \caption{Comparison with recent outstanding works across all metrics.}
    \label{fig:pentagon_comparison with baseline}
\end{figure}

To provide a more intuitive comparison of the experimental results, we have integrated all evaluation metrics from the experiments into a single graph. We did not include efficiency measures because the time differences between different models are significantly large, making it challenging to discern more granular differences in the same graph.
Among the above five metrics, Jaccard, F1-score, and PRAUC are accuracy tests, and DDI and Avg.\# of Drugs are safety tests. We unified multiple metrics based on their scores in various models and standardized them to a range between 0.3 and 0.9, and the resulting comparison is depicted in Figure \ref{fig:pentagon_comparison with baseline}. Given that lower values of DDI and Avg.\# of Drugs signify enhanced safety, their inverses are utilized for representation in Figure \ref{fig:pentagon_comparison with baseline}. Each color is associated with a particular model's outcome in the depicted pentagon. Each vertex represents the performance of a particular model on that metric, with vertices closer to the outer ring indicating superior performance on that metric.

GAMENet incorporates the patients' history, achieving enhanced accuracy compared to earlier models. Also, the model structure is relatively simple and moderately efficient. Nevertheless, there is room for improvement concerning the DDI rate. 
SafeDrug gets excellent DDI rates by considering DDI at the molecular level, and its limitation is not fully addressing the intricate association between medication and disease, which could further improve its precision. 
MICRON has significantly enhanced precision levels by utilizing an innovative approach that adjusts medication combinations based on a patient's health status changes between consecutive visits. Also noteworthy is the article's low complexity and pre-screening of the data, which makes it very fast in training.
COGNet, through its introduction of a translation model and a replication mechanism, has shown significant improvement in Jaccard. At the same time, because the model uses a small-batch training method, it requires many iterations to converge to the optimal result, which results in some lacking in time performance. Although it provides some measures to address safety, its core mechanism seems insufficient in fully representing safety concerns.
By integrating external molecular-based knowledge, MoleRec establishes a link between the substructure of molecules and disease. This approach has led to stable improvements in various accuracy metrics without significantly compromising safety. But due to its very large network and deep message passing strategy, it lags far behind other models in terms of training speed.

Our model globally explores entity associations, mitigating the long-tail distribution of medical data. By expressing the relationships between medications at the same level as well as the relationships between medications and diseases, we find a balanced point between safety and accuracy. 
At the same time, the performance in terms of efficiency remains outstanding due to the design of a concise and effective graph network structure and the application of pre-training methods.
Evaluated from both individual metrics and a comprehensive perspective, the performance of StratMed is significantly superior to other baselines.
Using MoleRec as the sub-optimal baseline model, comparing the above metrics, our model reduces safety risk (DDI rate) by 15.08\%, improves accuracy (Jaccard, F1-score, PRAUC) by 0.36\%, and reduces time expense (training time) by 81.66\%.

\section{Discussions}

In this section, we analyze the experimental results in the previous section in depth to highlight the advancement of our frame and conduct a large number of auxiliary experiments to verify the completeness and rationality of our work.

\subsection{Result Analysis}

The above results show that our model has some improvement in effectiveness and efficiency, so we launch an in-depth analysis of the experimental results.

\subsubsection{Effectiveness Analysis}

The data in medical datasets exhibits a high degree of imbalance, with most medications and diseases represented as sparse data. However, our baseline model has overlooked the importance of further exploration of such data. In contrast, our approach introduces a stratification module that categorizes the entire dataset into multiple levels based on the frequency of occurrence. Depending on the varying sparsity levels at each layer, we strengthen the expressive capacity of mid to low-level data with varying degrees of amplification. This approach not only allows us to continue thoroughly learning from popular data but also ensures the completeness of features learned for less common data, thus enhancing the representative capacity of all data. This method fills the gap that other models still need to consider, thereby providing a more effective solution to the long-tail issue in medical datasets and improving overall model performance.

As previous work \cite{molerec} has pointed out the problem,  the medication recommendation field faces the challenge of balancing safety and accuracy, where most models prioritize accuracy while safety receives insufficient attention. However, the presence of drug-drug interactions (DDIs) makes safety a critical concern specific to medication recommendations. Our model, featuring a dual-graph network structure, adeptly captures medication relationships for safety and medication-disease relationships for accuracy. This balanced approach enables simultaneous and unbiased expression of both safety and accuracy, resulting in substantial performance superiority over other models in comprehensive evaluations.

\subsubsection{Computational Complexity Analysis}

We first disassemble the model and analyze its complexity.
Our model comprises a pre-training module, a stratification module, a dual-property representation module, and a recommendation module.
First, for the pre-training module, which consists of a single linear layer, the time complexity is denoted as \(O({dim}^2)\), where \({dim}\) represents the embedding dimension of each medical entity.
Then, in the stratification phase, our model relies on statistical and rule-based computations that occur only before model training and do not participate in training. The time complexity of this phase is \(O(N)\), where \(N\) corresponds to the number of all visits.
The dual-property representation phase involves multiple Graph Convolutional Networks (GCN) and linear layer networks, leading to relatively higher complexity. The time complexity of the GCN component is \(O(n^2 \cdot {dim})\), and the time complexity of the linear layer is \(O({dim}^2)\). Consequently, the total time complexity for this module is \(O(n^2 \cdot {dim} + {dim}^2)\), where \(n\) signifies the number of medical entities in each visit.
Finally, the medication recommendation phase comprises multiple Gated Recurrent Unit (GRU) networks and linear layer networks, adding a certain degree of complexity. The time complexity of GRUs is \(O(T \cdot {dim}^2)\), while the linear layers exhibit a time complexity of \(O({dim}^2)\) as the previous phase. The overall time complexity for this module is \(O(T \cdot {dim}^2 + {dim}^2)\), with \(T\) denoting the time steps employed for each GRU. In summary, the time complexity of the primary model can be expressed as:
\[
O((({dim}^2)i)+N+(((n^2+(T+2){dim}){dim})j)),
\]
where $i$ is the hypothesized pre-training convergence step, and $j$ is the hypothesized master model convergence step.

In our time expenditure analysis, we found that models with high time consumption often have complex graph neural networks with multiple layers (3 to 5 layers). In contrast, our unique single-layer graph network construction approach keeps our time consumption stable during single-epoch training. Our stratification method, based on statistical principles and rule-based techniques, minimizes computational overhead. Additionally, our pre-training model uses minimal time, containing only data embeddings. With essential parameters pre-trained, our model converges quickly, outperforming most baseline models.

\subsection{Ablation Study}

\begin{table*}[width=.9\textwidth,cols=4,pos=h]
  \caption{The performance of the ablation study on the MIMIC-III test set in terms of accuracy and safety was evaluated using five evaluation metrics: Jaccard, DDI rate, F1-score, PRAUC, and Average number of drugs. The best and the runner-up results are highlighted in bold and underlined respectively under t-tests, at the level of 95\% confidence level.}
  \begin{tabular*}{\tblwidth}{@{} LLLLLL@{} }
   \toprule
    method & Jaccard$\uparrow$  & DDI$\downarrow $ & F1-score$\uparrow$  & PRAUC$\uparrow$  & Avg.$\#$ of Drugs \\
   \midrule
    StratMed $w/o$ P &	0.5055 $\pm$ 0.0020 & \textbf{0.0621 $\pm$ 0.0004} & 0.6625 $\pm$ 0.0018 & 0.7538 $\pm$ 0.0027	& 22.1602 $\pm$ 0.1675\\
    StratMed $w/o$ S & 0.5231 $\pm$ 0.0024 & \underline{0.0624 $\pm$ 0.0005} & 0.6774 $\pm$ 0.0022 & 0.7638 $\pm$ 0.0026 & 20.6739 $\pm$ 0.1369\\
    StratMed $w/o$ S+G & \underline{0.5279 $\pm$ 0.0026} & 0.0651 $\pm$ 0.0005 & \underline{0.6824 $\pm$ 0.0023} & \underline{0.7743 $\pm$ 0.0023} & 20.2318 $\pm$ 0.1523\\
    \midrule
    StratMed & \textbf{0.5321 $\pm$ 0.0035}  & 0.0642 $\pm$ 0.0005 & \textbf{0.6861 $\pm$ 0.0034} & \textbf{0.7779 $\pm$ 0.0043} & 20.5318 $\pm$ 0.1681\\
   \bottomrule
  \end{tabular*}
  \label{tab:ablation study}
\end{table*}

To further validate the effectiveness of each component we proposed, we also designed some simplified variants of our models. The primary study is to investigate the effect of pre-training, relevance stratification, and dual-graph network application and analyze its practical significance.

\textbf{StratMed $w/o$ P}:
We remove the pre-training module, and the initialization of entity embeddings changes from being provided by the pre-training module to being obtained by uniform distribution at the beginning of training.

\textbf{StratMed $w/o$ S}:
We removed the relevance stratification module, no longer balanced the long-tail distribution, and the sparse entity relationships do not have an enhancement mechanism. 

\textbf{StratMed $w/o$ S+G}:
Due to the topological structure between the graph network (dual-property representation) module and the stratification module, we cannot remove the graph network independently, so we removed the graph network module on the basis of removing the stratification module. This model will not consider the relationship between entities, directly merge biomedical entities into visit representations, and then recommend through the medication recommendation module.

Table \ref{tab:ablation study} shows the results for different variants of StratMed, whose phenomena are consistent with our ideas. StratMed $w/o$ P shows that pre-training entity embeddings are very effective, and more stable entity embeddings can enhance the expressiveness of the model. StratMed $w/o$ S shows the significance of our proposed relevance stratification module. By reinforcing the rare relationships of entities, it alleviates the disparity in item density brought about by the long-tail distribution. This process harmonizes the expression intensity between popular and sparse entities, enhancing the model's performance. Compared to the method that only removes the stratification module, the accuracy of StratMed $w/o$ S+G  increases but the safety decreases, and the relationship between the two is gradually tilted, which shows that our -graph network module has a crucial role in balancing the two properties.
However, there's still a decline when compared to the complete model. At the same time, it can be analyzed that the graph network module exhibits significant synergy with the relevance stratification module, efficiently capitalizing on the processed relationships. Their combined presence is essential to improve model performance.

In summary, the three modules we proposed are essential and have contributed to enhancing the model's performance to a certain extent.

\subsection{Parameter Sensitivity}

\begin{figure}
    \centering
    \includegraphics[width = \linewidth]{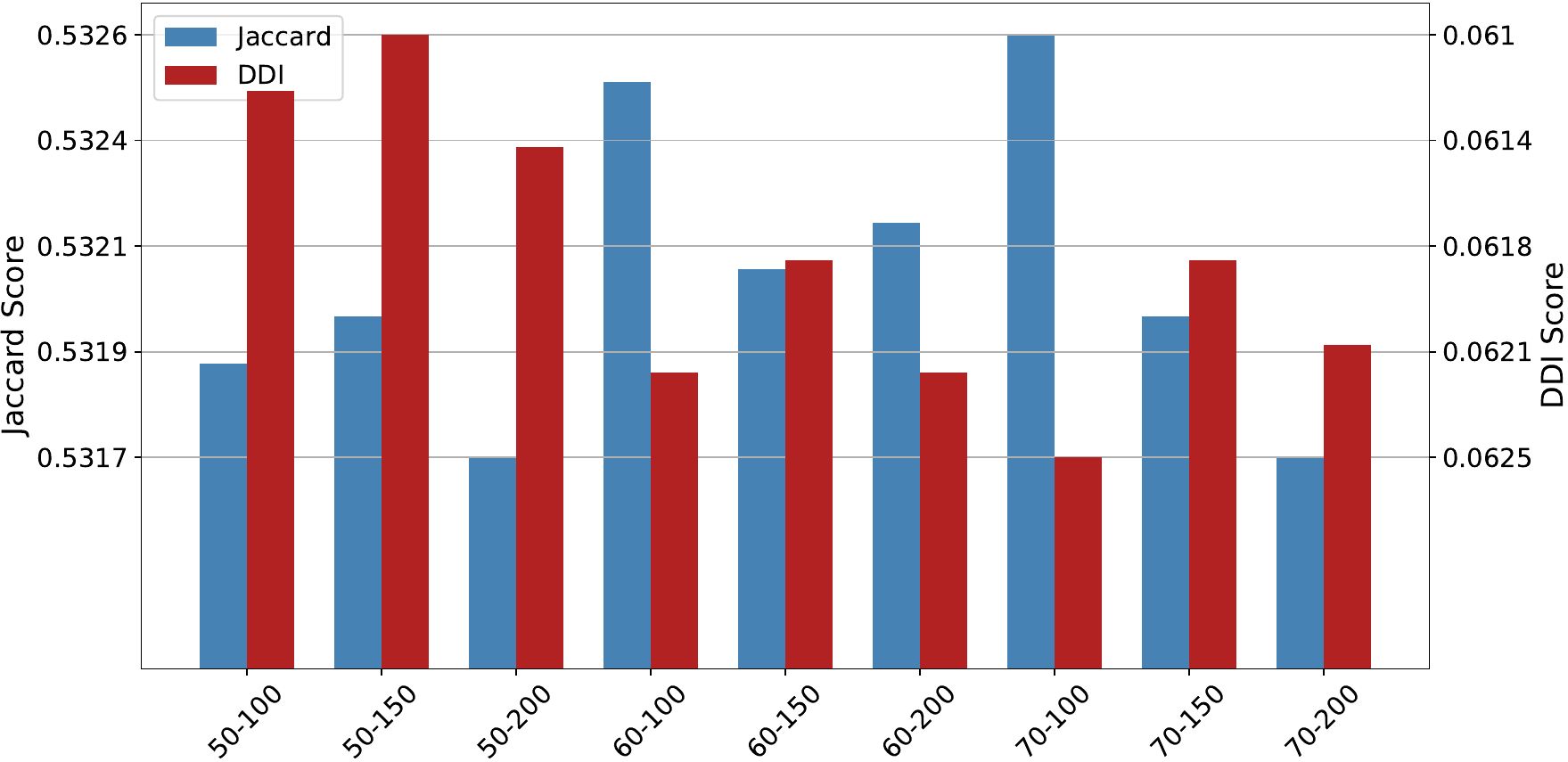}
    \caption{Performance of StratMed under varying stratification parameters, where taller bar indicates better model performance under that specific parameter combination.}
    \label{fig:parameter_sensitivity}
\end{figure}

Since the associations of each entity are captured under our stratification strategy, the gradient coefficients of the stratification and the amount of data in the top layer may impact the final results. Six sets of proposals are compared as Figure \ref{fig:parameter_sensitivity} to explore which parameter combinations yield better recommendation results:

We set the gradient coefficient $k$ = 2, after which we set the amount of data in the top level of the medication safety relevance $q_{m-m}$ = 50, 60, 70 and the amount of data in the top level of the two heterogeneous mapping relevance $q_{m-d}$ = $q_{m-p}$ = 100, 150, 200, respectively. Figure \ref{fig:parameter_sensitivity} shows the performance of StratMed in different stratification parameters, where the horizontal axis represents pairs of parameter combinations by \( q_{m-m} \) / \( q_{m-d} \), \( q_{m-p} \) and the blue bars represent precision (measured by Jaccard) and are mapped to the left y-axis. In contrast, the red bars denote safety (measured by DDI) and correspond to the right y-axis. Notably, a lower DDI indicates better safety. Consequently, the data on the right y-axis representing DDI is inverted. In this visualization, for both the blue and red bars, a greater height signifies superior model performance for both the blue and red bars. 

As shown in Figure \ref{fig:parameter_sensitivity}, since the model learns safety before accuracy, the changes in \( q_{m-d} \) and \( q_{m-p} \) generate fewer effects compared to variations in \( q_{m-m} \). When setting the top layer count \( q_{m-m} \) for medication safety stratification to 50, the safety improvement is more pronounced because a more minor top layer count results in more layers, enabling finer delineations of safety levels, thereby enhancing the ability to express variations in relationships, but this comes at a slight compromise in accuracy. 
Other experimental results also indicate the same trend, if the top layer count is set too high, it results in fewer layers, and numerous distinct relationships are categorized into the same class, making it challenging to express the differences between them. On the other hand, if set too low, each layer may lack sufficient samples, hindering the model's ability to be adequately trained and to capture a general representation for that layer.  Ultimately, to find the balance between safety and precision, we selected the parameter set with \( q_{m-m}/q_{m-d}, q_{m-p} \) = 60/150.

\subsection{Qualitative Study}
To investigate whether the stratification approach proposed in this work can adapt to imbalanced data environments and fundamentally address the issue of long-tail distributions, we conducted a series of relevant qualitative experiments.

\subsubsection{Overfitting Study}
Training on highly imbalanced datasets typically leads to models learning relatively common data more frequently and underlearning rare data, which can lead to overfitting problems for high-frequency data. As a result, the model may only perform well for specific data distributions and lack generalization ability in the face of other data distributions. Therefore, we conducted a series of experiments to investigate whether our proposed stratification approach can mitigate the effects of overlearning on specific data due to data imbalance, especially the overfitting performance in the presence of different degrees of data imbalance.

\begin{figure}
    \raggedright
    \includegraphics[width=0.9\linewidth]{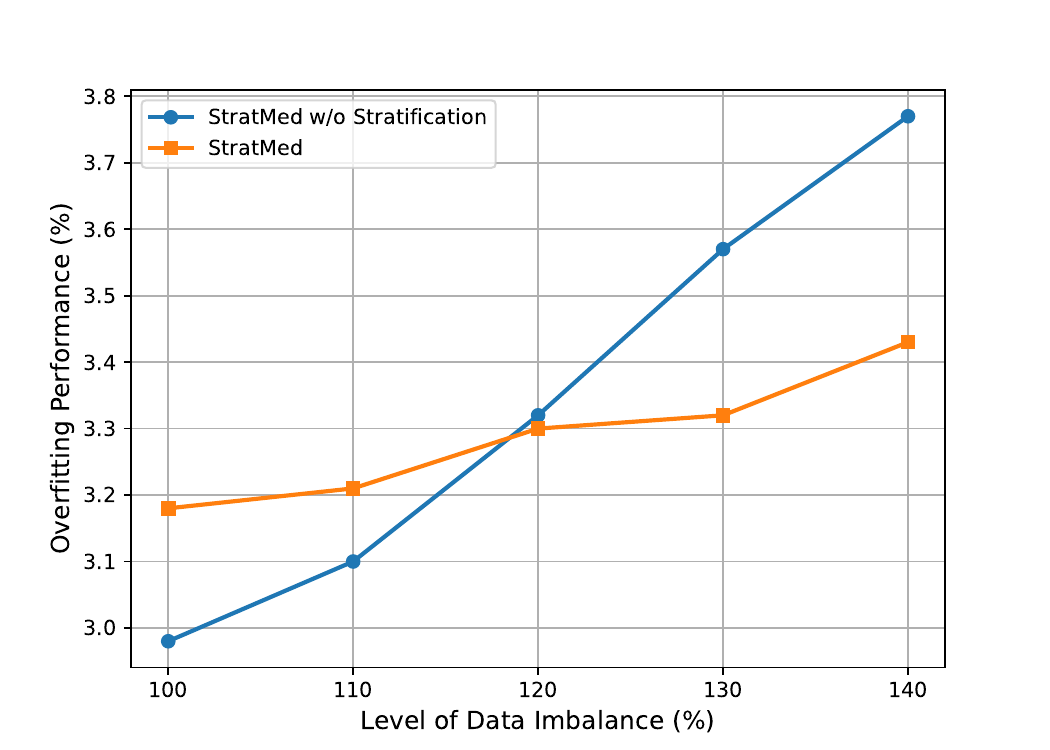}
    \caption{Overfitting performance with and without stratification algorithms in environments with varying levels of data imbalance.}
    \label{fig:overfitting}
\end{figure}

The data imbalance is characterized by some data being excessively high-frequency, while other data is excessively low-frequency. The severity of this imbalance is amplified as the data distribution becomes more extreme. To investigate the issue raised in the preceding text, we initially created multiple data environments with varying degrees of distortion. We categorized the data into three levels based on their occurrence frequency: common, moderate, and rare. From the global dataset, we removed the moderate-level data, retaining only the common and rare data, to construct a more extreme data context. We adjusted the degree of data distortion by controlling the amount of removal of the moderate-level data. Considering that medical data often exhibits inherent distortion, we defined the untreated data as having 100\% distortion, and then generated data with distortion levels ranging from 110\% to 140\% by removing 10\% to 40\% of the middle-level data volume. Subsequently, we conducted evaluations of the complete StratMed model and the StratMed model without the stratification module in those five different levels of data environments to compare the impact of stratification on mitigating overfitting issues in imbalanced data. It is worth noting that we were unable to exacerbate the imbalance levels to 150\% as, at that point, the model became non-operational.

As illustrated in Figure \ref{fig:overfitting}, without the assistance of a stratification algorithm, the model's overfitting becomes increasingly severe as the data imbalance grows. This phenomenon is particularly pronounced when the imbalance degree reaches 140\%. This is primarily because, during training, it becomes nearly impossible to capture the features of low-frequency data, resulting in a heavy reliance on high-frequency data. Consequently, when the data distribution exhibits subtle variations, the learned features struggle to adapt, leading to a significant loss in accuracy and pronounced overfitting issues. In contrast, after the application of the stratification algorithm, although the overfitting phenomenon can not be eliminated, this problem is improved. When the data becomes more and more unbalanced, the stratification algorithm will give more care to the low-frequency data at the bottom, continuously amplifying the learning effect of the bottom layer, and correcting the learning bias can also be achieved based on the algorithms without using traditional algorithms against long-tailed distributions, such as oversampling. This demonstrates the ability of our model to mitigate the effects of over-learning on specific data due to data imbalance.

\subsubsection{Robustness Study}

The most crucial challenge in addressing long-tail distribution lies in enhancing the learning capabilities for sparse data. To substantiate the effectiveness of the stratification algorithm proposed in this paper in improving sparse data learning, we conducted robustness experiments on various prominent models across different levels of data sparsity.

We reprocessed the dataset to classify the low-frequency scenarios of the dataset according to the frequency of occurrence of diagnosis/procedure in all visits. We set the low-frequency thresholds $\mu$ to have 5\%, 10\%, 15\%, and 20\%, e.g., at $\mu$ =10\%, all high/middle-frequency entities are erased, and only those diagnoses and procedures with a frequency of occurrence less than 10\% are retained.

\begin{figure}
    \raggedright
    \includegraphics[width=0.9\linewidth]{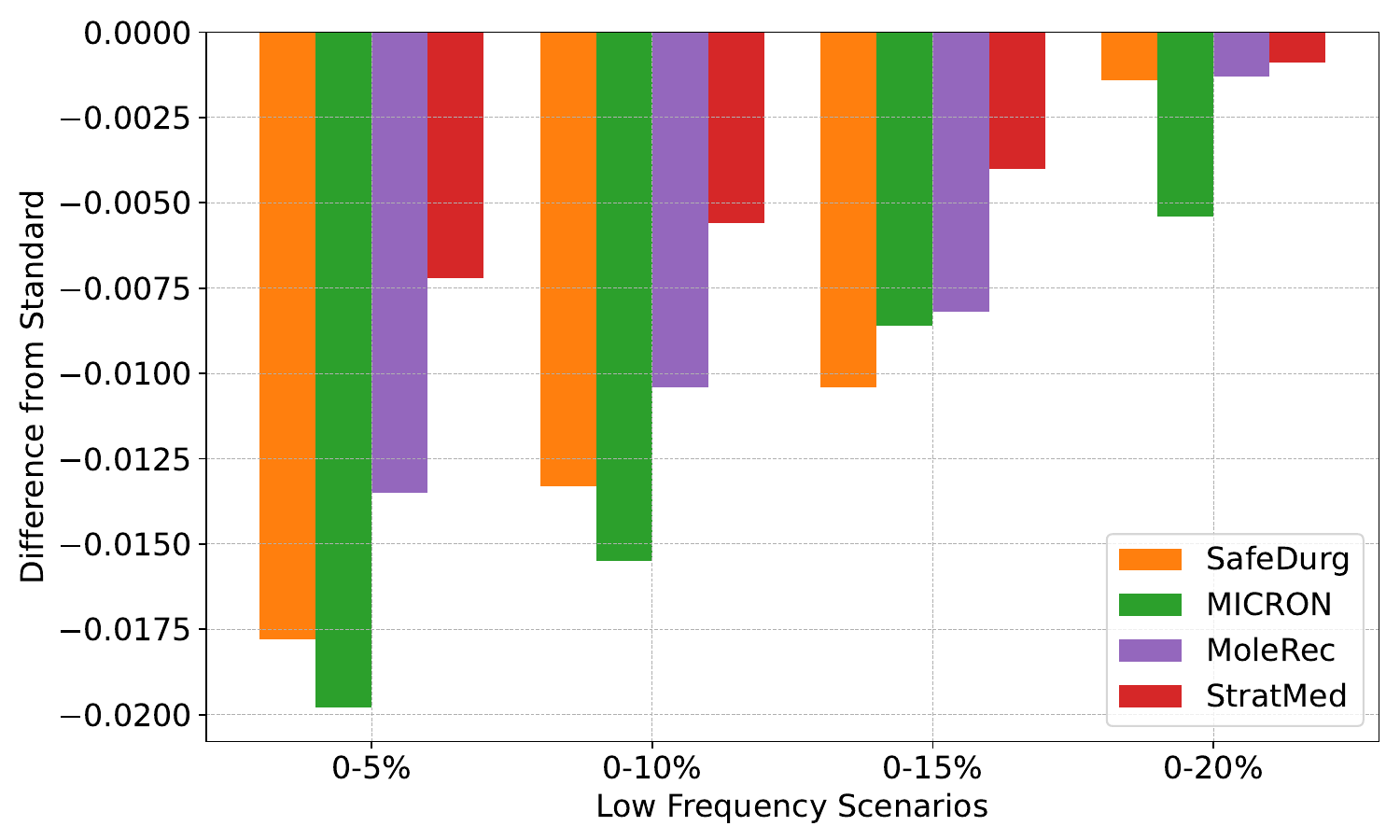}
    \caption{Performance of different models under sparse scenarios, where taller bar indicates a serious impact on the model under that specific scenario.}
    \label{fig:low_frequency}
\end{figure}

We have compared our model with state-of-the-art models from recent years, and the results are illustrated. In Figure \ref{fig:low_frequency}, the horizontal axis represents different degrees of low-frequency data, while the vertical axis depicts the difference in the performance of the model under sparse conditions compared to its performance on the original dataset (where the Jaccard metric represents precision). 
The histogram indicates the extent to which each model is affected. (GAMENet and COGnet were disproportionately affected to the point that they couldn't provide complete recommendations, leading to their exclusion from the figure.) 
The fluctuations are most pronounced when the sparse severity is high. However, when the severity of sparsity eases into the 0-20\% range, the impact is minimal, with some models being virtually unaffected. Specifically, both SafeDrug and MICRON experience a notable impact. MoleRec is less affected due to its incorporation of external molecular structures, ensuring the connection between medications and diagnoses/procedures. In contrast, our model maintains stable performance even under sparse data, attributable to its adaptive stratification method that enhances sparse relationships, thereby exhibiting robustness against data sparsity.

\subsection{Case Study}
We randomly selected a prescription process from one of the visits in the MIMIC-III test dataset as a demonstration of how our method recommends medications. For clarity in presentation, we chose a relatively uncomplicated medical record with more diagnosis items and fewer procedure items, precisely 12 diagnoses, and a single procedure. Since the only given procedure \#706 is a rare entity, our subsequent analysis primarily focuses on the relevance between diagnoses and medications.

\begin{figure}
    \centering
    \begin{subfigure}{\linewidth}
        \raggedright
        \includegraphics[width=0.9\textwidth]{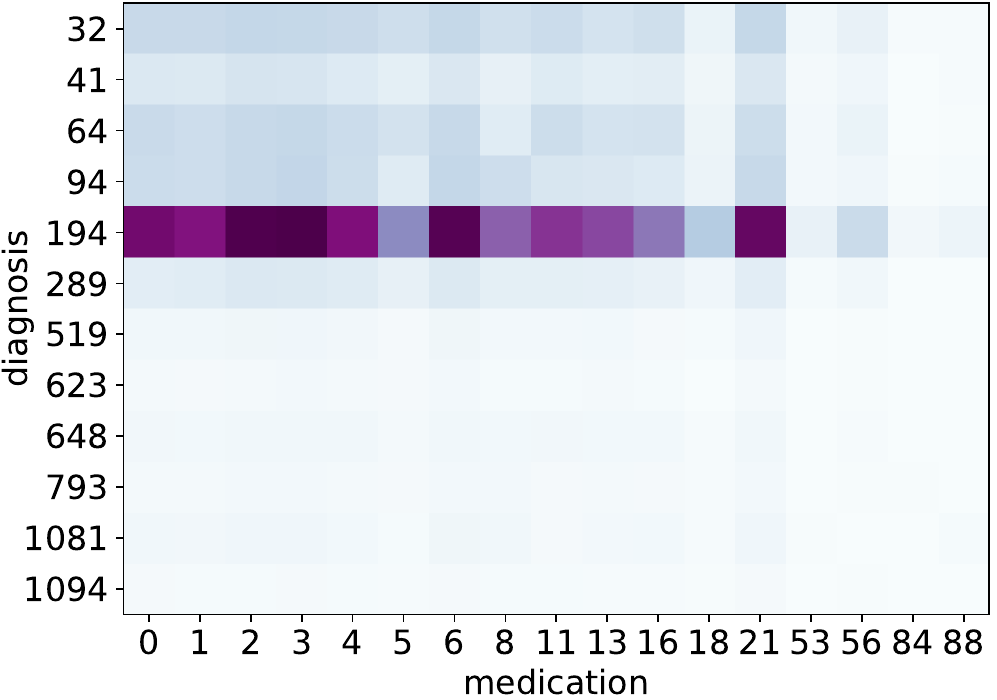}
        \caption{relevance before stratification}
        \label{fig:diag_med_relevanceA}
    \end{subfigure}
    
    \vspace{0.2cm}  
    
    \begin{subfigure}{\linewidth}
        \raggedright
        \includegraphics[width=0.9\textwidth]{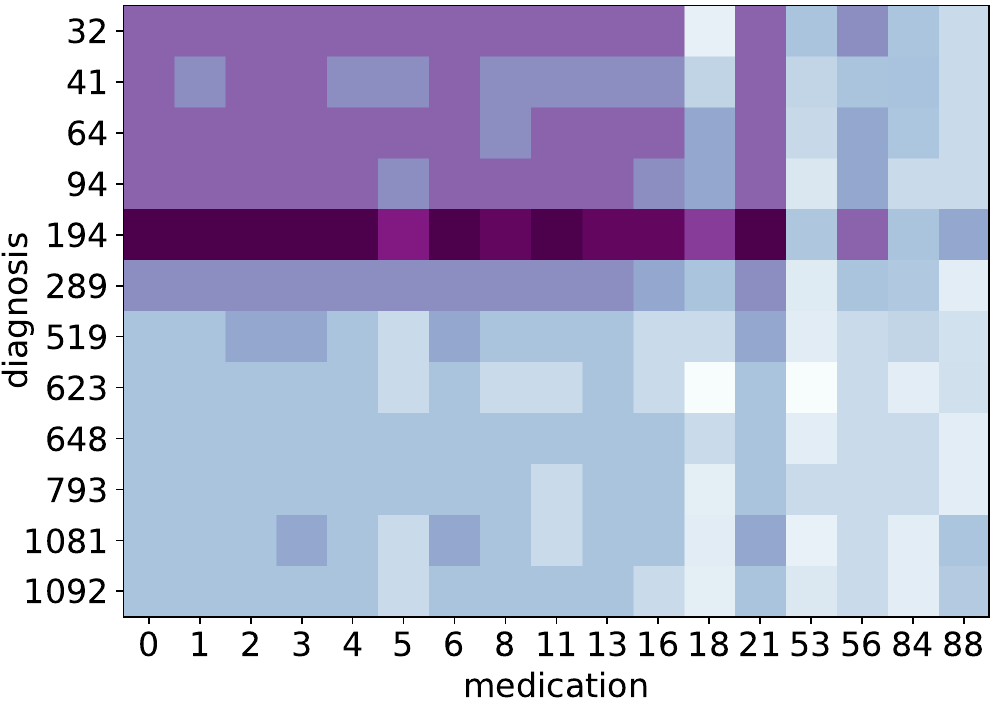}
        \caption{relevance after stratification}
        \label{fig:diag_med_relevanceB}
    \end{subfigure}
    
    \begin{subfigure}{\linewidth}
        \centering
        \includegraphics[width=0.9\textwidth]{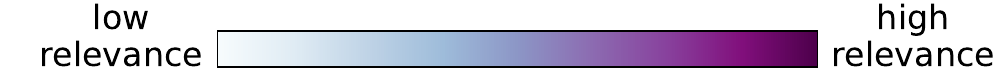} 
    \end{subfigure}
    
    \caption{Distribution comparison of diagnosis-medication relevance before and after relevance stratification, with the Figure \ref{fig:diag_med_relevanceA} representing before stratification and the Figure \ref{fig:diag_med_relevanceB} representing after stratification.}
    \label{fig:diag_med_relevance}
\end{figure}

\begin{figure}
    \centering
    \includegraphics[width=0.9\linewidth]{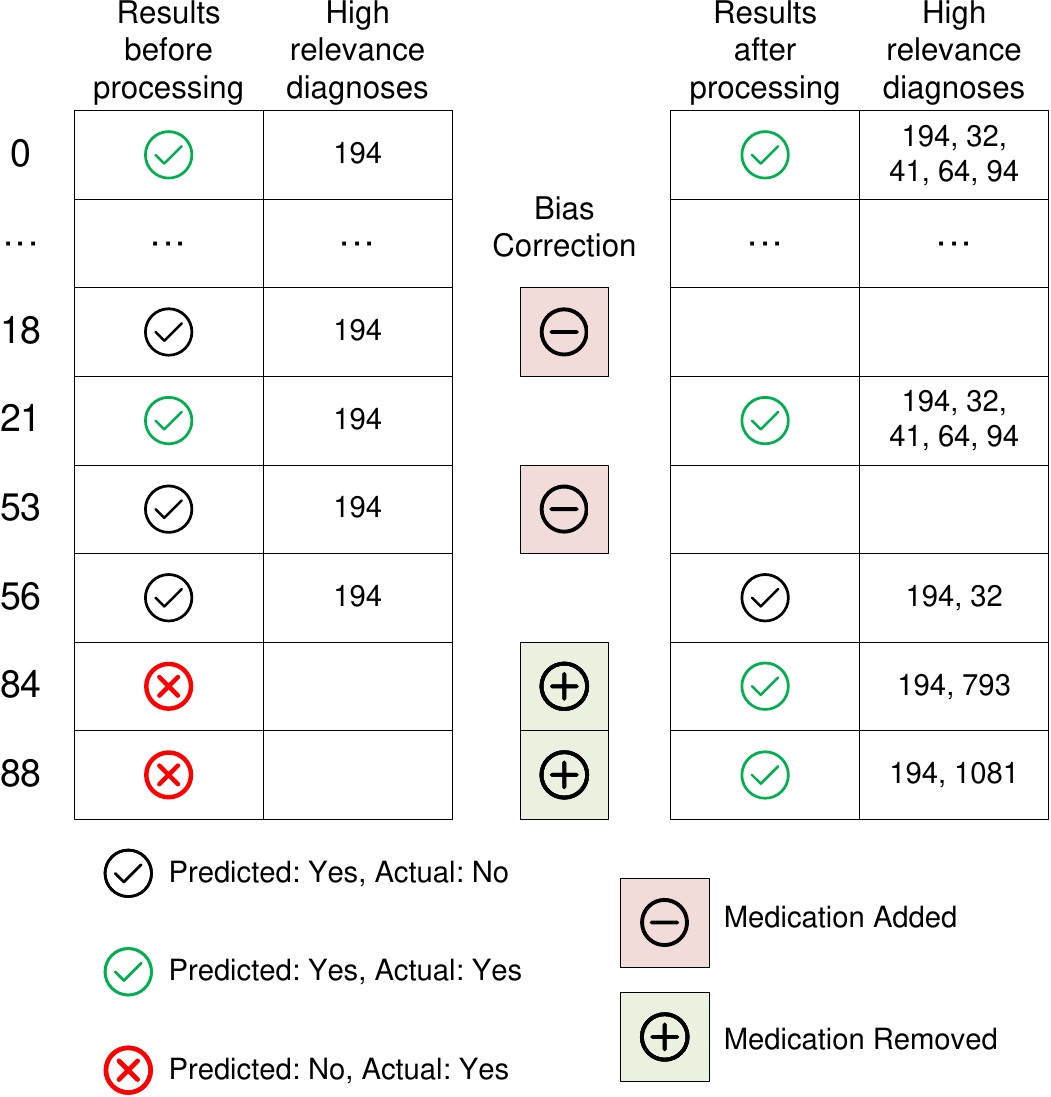}
    \caption{A visualization of the relevance stratification process during the medication prescribing. The left figure depicts the recommendation results before bias correction, while the right figure depicts the results after bias correction.}
    \label{fig:case_study}
\end{figure}

In Figure \ref{fig:diag_med_relevance} and \ref{fig:case_study}, we demonstrate the effect of bias correction arising from the relevance stratification phase. Figure \ref{fig:diag_med_relevance} portrays the distribution of diagnosis-medication relevance before and after applying the relevance stratification. On the y-axis, we denote the diagnoses from the current visit, whereas the x-axis denotes the union of predicted and actual medications. Also, Figure \ref{fig:case_study} shows the medication recommendations before and after bias correction and the main reason for prescribing this medication (the diagnoses that have a high relevance with this medication). In this illustration, the numbers on the left denote the union of predicted and actual medications, we only demonstrate the parts that have seen changes. Green checkmarks indicate medications present in both the recommended and actual prescriptions (correct prediction), black checkmarks indicate medications present in the recommended prescription but absent from the actual one (over prediction), and red crosses indicate medications absent in the recommended prescription but present in the actual one (error prediction). During bias correction, pink minus signs operation the removal of medications, and green plus signs operation the addition of medications.

From the figures, it is evident that before correction, the relevance between diagnosis \#197 and all the medications was dominant, resulting in all recommended medications being sourced from this diagnosis. This imbalanced distribution amplified the representation of specific diagnoses while limiting the model's capability to represent others. Our method effectively balances the distribution. In the bias correction phase, we found that medications \#18 and \#53, whose associations with diagnosis \#197 were over-expressed, were redundant medications for that diagnosis and were therefore excluded. At the same time, medications \#84 and \#88 were added to the medication recommendations because the model overlooked their association with the diagnoses due to their sparsity.

Post-correction results showcase a noticeable enhancement in accuracy, underlining the significance of the relevance stratification strategy we introduced.

\section{Conclusion}

In this research, we introduce a model for medication recommendation named "StratMed." This method incorporates a novel frequency-dependent stratification approach, significantly enhancing the learning capacity for low-frequency data and effectively mitigating the challenges associated with the long-tail problem in medical datasets. Furthermore, our study employs a dual-property representation frame, which expresses both the safety and accuracy aspects of medication combinations. This innovative approach alleviates the mutual constraints between these two essential properties. Our research includes rigorous experiments conducted on publicly available clinical datasets. The experimental findings demonstrate that our approach excels in three critical aspects: accuracy, safety, and efficiency.

Although our research has made great strides in medication recommendation, we have some limitations that need to be further explored. As new medications and diseases continue to emerge, existing medication recommendation methods often struggle to make recommendations for data not present in the training dataset. In the future, based on StratMed, we will pay more attention to solving the "cold-start" problem so that the model can be better adapted to real-world scenarios.

\printcredits
\section*{Declaration of competing interest}
The authors declare that they have no known competing financial interests or personal relationships that could have appeared to influence the work reported in this paper.

\section*{Acknowledgments}
\subsection*{Funding}
This work was supported by the Innovation Capability Improvement Plan Project of Hebei Province (No. 22567637H), the S\&T Program of Hebei(No. 236Z0302G), and HeBei Natural Science Foundation under Grant (No.G2021203010 \& No.F2021203038).

\bibliographystyle{cas-model2-names} 

\bibliography{StratMed}

\bio{}
\endbio

\bio{}
\endbio

\end{document}